\pdfoutput=1
\documentclass[10pt,twocolumn,letterpaper]{article}

\usepackage{iccv}              

\usepackage{graphicx}
\usepackage{amsmath}
\usepackage{amssymb}
\usepackage{booktabs}
\usepackage{adjustbox}
\usepackage{multirow}
\usepackage{arydshln}
\usepackage{colortbl}
\usepackage{amsfonts}       
\usepackage{nicefrac}       
\usepackage{microtype}      
\usepackage{times}
\usepackage{epsfig}
\usepackage{tipa}
\usepackage[T1, OT1]{fontenc}
\DeclareTextSymbolDefault{\dh}{T1}
\usepackage{booktabs}
\usepackage{multirow}
\usepackage{color, colortbl}
\usepackage{pifont}
\usepackage{makecell}
\usepackage{threeparttable}
\usepackage[shortlabels]{enumitem}
\usepackage{xcolor}
\usepackage{scalerel}
\usepackage{makecell}
\usepackage{tabu}
\usepackage{tikz}
\usepackage{marvosym}
\usepackage{xcolor} 

\definecolor{Gray}{gray}{0.5}
\definecolor{LGray}{gray}{0.9}
\definecolor{darkblue}{RGB}{94,110,186}
\definecolor{darkGreen}{RGB}{92, 148, 110}
\definecolor{myblue}{RGB}{14, 121, 178}
\definecolor{rightpath}{RGB}{248, 203, 173}
\definecolor{wrongpath}{RGB}{89, 89, 89}

\newcommand{\darkGreen}[1]{\textcolor{darkGreen}{#1}}

\newcommand{\modelname}{VideoChat-Flash}
\def\methodname{hierarchical compression}
\def\shortname{HiCo}
\def\sysname{VideoChat-Flash}

\definecolor{iccvblue}{rgb}{0.21,0.49,0.74}
\usepackage[pagebackref,breaklinks,colorlinks,allcolors=iccvblue]{hyperref}

\def\paperID{2104} 
\def\confName{ICCV}
\def\confYear{2025}

\title{VideoChat-Flash: Hierarchical Compression for Long-Context Video Modeling}

\author{
    Xinhao Li$^{2,1*}$, 
    Yi Wang$^{1,4*\dagger}$,
    Jiashuo Yu$^{1*}$,
    Xiangyu Zeng$^{2,1}$,
    Yuhan Zhu$^{2}$,\\
    Haian Huang$^{1}$, Jianfei Gao$^{1}$, Kunchang Li$^{3}$, Yinan He$^{1}$, Chenting Wang$^{1}$ \\
    Yu Qiao$^{1}$, Yali Wang$^{3,1}$, Limin Wang$^{2,1\dagger}$ \\
    \small$^1$Shanghai AI Laboratory~~~
    \small$^2$Nanjing University \\
    \small$^3$Shenzhen Institutes of Advanced Technology, Chinese Academy of Sciences \\
    \small$^4$Shanghai Innovation Institute \\
    {\small \url{https://github.com/OpenGVLab/VideoChat-Flash}}
}

\begin{document}
\maketitle
\begin{abstract}
Long-context video modeling is critical for multimodal large language models (MLLMs), enabling them to process movies, online video streams, and so on. Despite its advances, handling long videos remains challenging due to the difficulty in efficiently understanding the extremely long video context. This paper aims to address this issue from aspects of the model architecture, training data, training strategy and evaluation benchmark. First, we propose a novel \textbf{Hi}erarchical video token \textbf{Co}mpression (\textbf{HiCo}) method, which leverages visual redundancy in long videos to compress long video context from Clip-level to Video-level, reducing the computation significantly while preserving essential details, achieving an extreme compression ratio of approximately \textbf{1/50} with almost no performance loss. Second, we introduce a multi-stage \textbf{short-to-long learning} scheme, a large-scale dataset of real-world long videos named \textbf{LongVid}, and a challenging \textit{``Multi-Hop Needle-In-A-Video-Haystack''} benchmark. Finally, we build a powerful video MLLM named \textbf{\modelname{}}, which shows a leading performance on both mainstream long and short video benchmarks at the 2B and 7B model scale. It first gets \textbf{99.1\%} accuracy over 10,000 frames in NIAH among open-source models.

\end{abstract} 
{
\renewcommand{\thefootnote}%
{\fnsymbol{footnote}}
\footnotetext[0]{* Equal contribution. $\dagger$ Corresponding authors.} 
}
\section{Introduction}
Long-context video modeling stands as one of the most crucial capabilities within multimodal large language models (MLLMs). This capability empowers MLLMs to proficiently manage hours-long movies, documentaries, and online video streams, all of which demand sophisticated long video processing.
\begin{figure}[!ht]
    \centering
    \includegraphics[width=\linewidth]{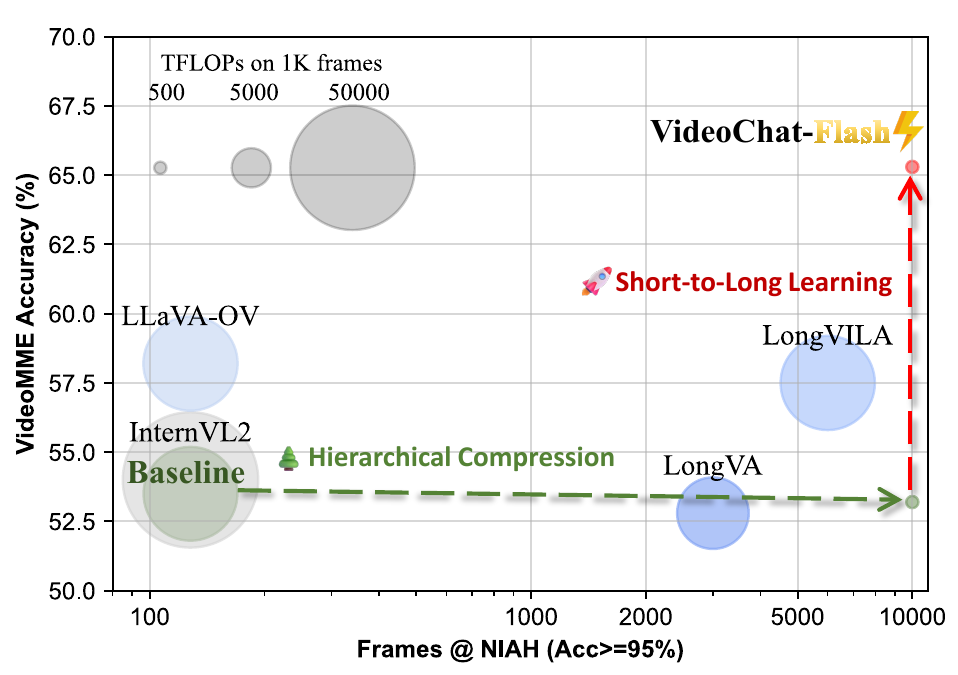}
    \caption{\textbf{Comparison with mainstream MLLMs for long videos.} \modelname{} improves long video understanding efficiency and effectiveness by hierarchical compression and a short-to-long learning approach, respectively.}
    \label{fig:teaser}
    \vspace{-4mm}
\end{figure} Recent advances in MLLMs~\cite{internvideo,internvideo2,li2023videochat,videochat2,videollama1,videollama2,llavanextvideo,videollava,pllava,llavaonevision,fuyu,otterhd} perform well in short video understanding. However, it remains challenging to build MLLMs for processing extremely long videos (lasting for hours or even longer). The difficulty lies in how to enable MLLMs to efficiently understand the extremely long video context brought by long videos.

\begin{figure*}
    \centering
    \includegraphics[width=0.96\linewidth]{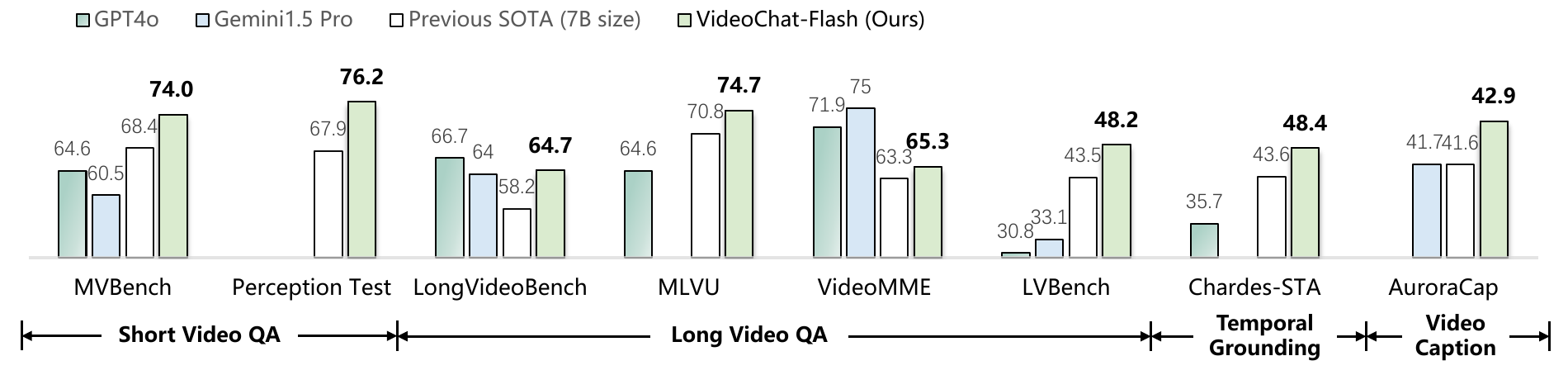}
    \caption{\textbf{Comparison results on various generic video-linguistic tasks}}
    \label{fig:sotafig}
\end{figure*}

Inspired by large language models (LLMs) with long context, modeling multimodal long context is widely studied from several perspectives. Some work~\cite{gemini,longvila} represented by Gemini-1.5-Pro~\cite{gemini} address it by training well-performed MLLMs on long-form corpus e.g. lengthy text and thousands of frames from videos, minimizing the gap between the evaluation and learning. Although the progress in system construction and hardware has made it possible to train and infer with super-long multimodal contexts, such super-long multimodal contexts have significantly reduced the training and inference efficiency of models. (For Gemini-1.5-Pro~\cite{gemini}, a one-hour video will be converted into 921,600 tokens). Meanwhile, the high redundancy in long video context makes it particularly difficult for models to understand. Some previous efforts~\cite{moviechat,llamavid,longvu} have been made to compress video tokens in order to achieve higher training and inference efficiency for long videos. However, the compression of visual content inevitably leads to the loss of detailed information. In some long video understanding benchmarks, certain current long video models even perform worse than some image-based MLLMs. Therefore, how to strike a balance between performance and efficiency remains a significant challenge. In this paper, we attempts to address the above issues from the model architecture, training data, training strategy and evaluation benchmark. 

First, we propose a novel \textbf{Hi}erarchical video token \textbf{Co}mpression method (\textbf{\shortname}) to model the long video context efficiently, which defines the compression of the long video context into two stages. First, we segment the long video into multiple clips. Then, at the Clip-level, we utilize the spatio-temporal attention of the video encoder and the similar token merging to aggregate the key information between frames, thereby reducing the redundancy of inter-frame features. Subsequently, we take advantage of the sparsity of attention when the LLM processes long video tokens, further discard the video tokens that are irrelevant to the current task at the Video-level. HiCo could achieve an extreme compression ratio of approximately 1/50 with almost no performance loss. Additionally, we have conducted thorough explorations of other designs such as video sampling and timestamp awareness prompt. 

Second, to further enrich the existing long video training corpus, we construct \textbf{LongVid}, a dataset that contains 300,000 hours of videos and 2 billion words of textual annotations. With LongVid, we have designed a multi-stage training strategy named \textbf{short-to-long learning}. The main idea is to first utilize image and short video data to learn basic visual perception abilities. Then, through the joint training of short video and long video data, the model is enabled to handle videos of different lengths and different types of tasks. In addition, we design a new evaluation benchmark named \textit{\textbf{``Multi-Hop Needle In A Video Haystack"}}. Which is more challenging and can better examine the model's complex reasoning abilities regarding long videos.

Finally, we develop a powerful video MLLM named \textbf{\sysname{}}, as shown in \cref{fig:sotafig}, which achieves remarkably leading performance with extremely high efficiency on various video understanding benchmarks. Even with a 7B size, it outperforms closed-source models such as GPT-4o~\cite{gpt4o} and Gemini-1.5-Pro~\cite{gemini}. And it first yields 99.1\% retrieval accuracy over 10,000 frames in the ``Needle-In-a-Video-Haystack'' among open-sourced MLLMs.

   
    
\section{Related Works}

\begin{figure*}[!ht]
    \centering
    \includegraphics[width=0.9\linewidth]{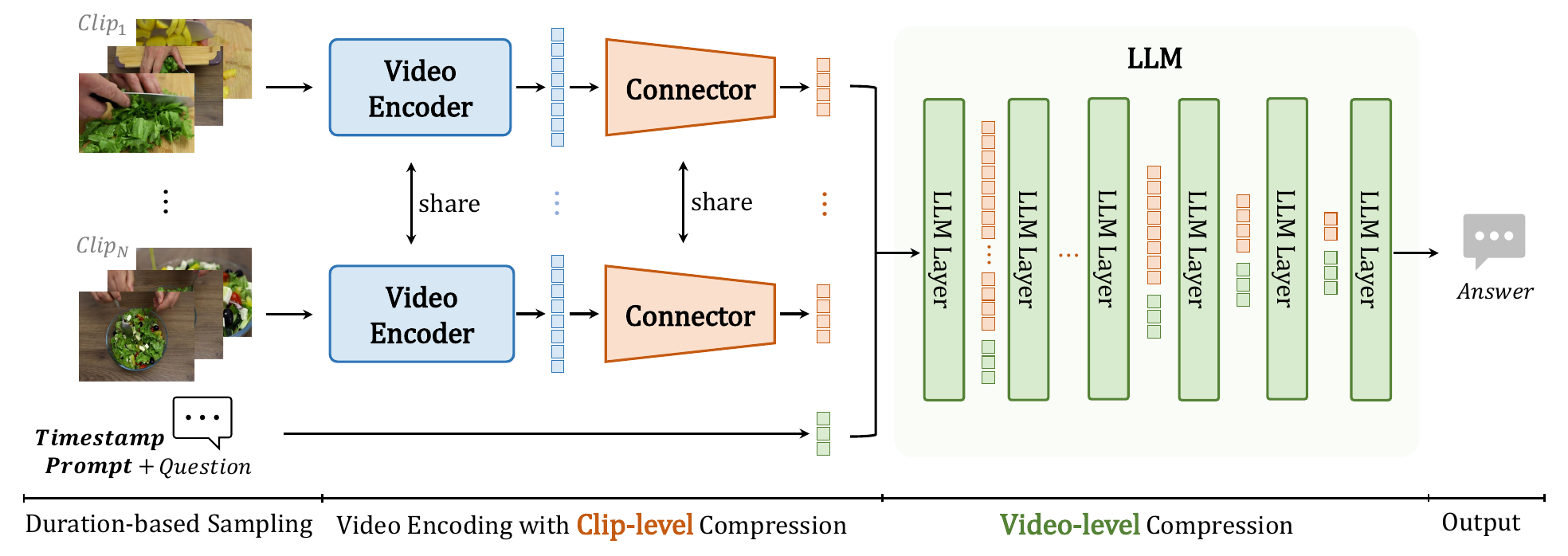}
    \caption{\textbf{Framework of \modelname{} with Hierarchical Video Token Compression.} Video tokens will be compressed at the Clip-level by leveraging the local redundancy of the video modality during video encoding. Subsequently, during LLM processing, they will be compressed at the Video-level by taking advantage of the sparsity in the interaction between the text modality and the video modality.}
    \label{fig:overview}
\end{figure*}

\paragraph{Multimodal Large Language Models for Video Understanding.} Recent advancements in multimodal large language models (MLLMs) have shown significant promise in video understanding. Most of them~\cite{li2023videochat,videochat2,internvideo2,videollava,videollama1,llavanextvideo,llavavideo} focus on the understanding of minute-level videos, and some works~\cite{gemini,moviechat,longllava,longvu,longvila,videoxl,videochatonline} have further tried to handle longer hour-level videos. To address the challenge of processing long videos, researchers focus on two key strategies: (1) extending the context window of the LLM~\cite{gemini,longva,longvila,uneven} and (2) compressing the video tokens~\cite{llamavid,videoccam,longvlm,koala,moviechat,videoxl,timesuite}. For context extension, although the approach of expanding the context window enable the possibility of long video understanding, it falls short of reducing the high computational burden and processing costs induced by long videos, thereby imposing limitations on its practical application. For token compression, Methods represented by Llama-Vid \cite{llamavid} use a highly compact representation while preserving key information. The high compression ratio makes it difficult for such methods to achieve excellent long video understanding performance, and they may even be inferior to some MLLMs designed for image modeling. Therefore, how to design a Video MLLMs architecture that can balance both efficiency and performance remains a difficult challenge. In this work, we provide a comprehensive solution that balances both efficiency and performance from various aspects such as the model architecture, training data, and training strategies.

\paragraph{Long Video Benchmark.} In order to evaluate the ability of Video MLLMs to understand long videos, previous works~\cite{moviechat,movqa,egoschema,cinepile,longvideobench,mlvu,videomme, hourvideo,lvbench} have achieved this by collecting long videos and then designing various multiple-choice questions related to the content of these long videos. This approach is closer to real-world applications and can effectively examine the model's ability to understand and reason about long videos. However, when it comes to examining the model's capabilities for videos of different lengths, this method is not intuitive enough. Inspired by the popular "Needle in A Haystack" (NIAH) evaluation in long text context evaluation, some recent works~\cite{longva,videoniah} have attempted NIAH for Video haystack. Nevertheless, it is difficult to assess complex reasoning abilities, and there may be information leakage. In this paper, we propose a more challenging \textit{``Multi-Hop Needle-In-A-Video-Haystack"} is designed to address the above issues.

\section{Method}

\subsection{\shortname: Efficient Long Video Modeling}
\label{sec3_1}

To enable MLLMs to handle thousands of input frames, we propose a new video context compression paradigm named \methodname{} (\shortname). This paradigm decomposes video context compression into two main stages: 1. \textbf{Clip-level} Compression during the encoding of long videos. 2. \textbf{Video-level} Compression within the context interaction in the LLM. Based on this framework, we have designed an innovative efficient Video MLLM architecture, VideoChat-Flash, as illustrated in \cref{fig:overview}. Below, we elaborate on our specific design details from data input to model output.

\paragraph{Duration-based Sampling.} First, we need to perform frame sampling on the original video. Specifically, we sample a raw video with a duration of $D$ to obtain $T$ frames as input. Considering that the requirements for understanding short and long videos often differ, we aim to conduct dense sampling on short videos to capture detailed motions and sparse sampling on long videos to focus on event understanding. To this end, we have designed a Duration-based Sampling strategy:
\begin{equation}
    T = \min(T_{\text{max}}, \max{(D, T_{\text{min}})}).
\end{equation}
Simultaneously, we define the sampling density $\phi$ as follows:
\begin{equation}
    \phi(T,D) = \frac{T}{D} = \frac{\min(T_{\text{max}}, \max{(D, T_{\text{min}})})}{D}.
\end{equation}
That is, for short videos where $D < T_{\text{min}}$, $\phi=T_{\text{min}}/D$ , which increases as the video length decreases. For long videos where $D > T_{\text{max}}$, $\phi=T_{\text{max}}/D$, which decreases as the video length increases.

\paragraph{Timestamp Prompt.} For video MLLMs, the ability to perceive timestamps is also a crucial capability. Unlike previous works~\cite{timechat,timesuite} that rely on additional modules or designs to achieve this (there is a considerable computational burden when there are a large number of video frames), we employ a simple timestamp prompt after the video context: \textbf{\textit{“The video lasts for N seconds, and T frames are uniformly sampled from it.”}} We find that this straightforward approach is sufficient to enable the model to perceive the timestamps of the input video, achieving excellent performance on timestamp sensitive tasks such as temporal grounding (see \cref{tab:main}).

\paragraph{Spatio-Temporal Compression Encoding for Clips.} Considering the substantial redundant and repetitive information, such as that of backgrounds and objects, present between adjacent frames in natural videos, we segment the original video frames into several clips. Subsequently, we employ a video encoder with spatio-temporal attention to encode these clips. This enables each visual token to aggregate information from other frame tokens as comprehensively as possible. Finally, we utilize token merging to combine highly similar tokens. Formally, given a frame sequence sampled from the original video, we divide it into $N_c$ equally sized clips. The frames of $j_{\text{th}}$ clip $\mathbf{x}^j$ are transformed by a video encoder and a connector $\mathcal{F}$, resulting in $M$ compressed visual tokens:
\begin{equation}
    [\mathbf{v}_i^j]_{i=1,2,..,M} = \mathcal{F}(\mathcal{V}(\mathbf{x}^j)),
\end{equation}
where $\mathcal{F}$ consists of a parameter-free similar token merge operation and an MLP projection. Ultimately, we concatenate the compressed tokens of each clip to obtain the input  for the LLM:
\begin{equation}
    \mathbf{X_v}= \text{Concat}([\mathbf{v}_i^1]_{i=1,2,..,M}, \cdot\cdot\cdot, [\mathbf{v}_i^{N_c}]_{i=1,2,..,M}).
\end{equation}

Benefiting from the effectiveness of the video encoder in modeling spatio-temporal interactions, we achieve an extremely heavy compression while well retaining the key information, with each video frame being compressed to an average of only \textbf{16} tokens.

\paragraph{Progressive Visual Dropout in LLM.} Although clip-level compression has been carried out before, due to the possibility of longer-range visual redundancies in long videos (e.g. surveillance videos), and when an LLM responds to specific instructions regarding the visual input, it may not be necessary to continuously focus on the entire long video context. We consider conducting further video-level compression during the LLM inference stage. Recent works~\cite{fastv,llavolta} have explored acceleration strategies for MLLMs when processing short visual contexts. Most of them drop visual tokens based on the correlation between text tokens and visual tokens. In contrast, we find that when the LLM processes a long video context, it pays attention to the entire long video context at the shallow layers of the LLM, while focusing on the details of certain local moments at the deep layers (see the Appendix for specific visualizations). Based on this observation, we have designed a progressive visual dropout strategy, which is divided into two stages. At the shallow layers of the LLM, we uniformly drop a small number of video tokens (i.e. uniform drop), reducing the computation while maintaining the original spatio-temporal structure of the video context. At the deep layers of the LLM, we rely on the correlation between text tokens and video tokens to retain the most critical relevant information (i.e. text-guided select). We have found that this operation not only effectively improves the computational efficiency of the model but also slightly enhances the understanding performance of the model by reducing irrelevant visual noise.


\subsection{Large-scale Corpus for Long Video Training} \label{sec:data}

One of the challenges in long video model training is the shortage of large-scale, high-quality data. Though recent advances have mitigated this by long-form datasets of video-text pairs, these lack the instruction-following paradigm, such as (video, instruction, answer) triplets, crucial for multimodal reasoning. To address this, we introduce a large-scale long video instruction-tuning dataset named \textbf{LongVid}. It comprises 114,228 long videos and 3,444,849 question-answering (QA) pairs across five different task types, supporting models to handle diverse long video scenarios.

To build LongVid, we leverages the rich diversity of existing datasets, including Ego4D~\cite{ego4d}, HowTo100M~\cite{howto100m}, HD-Vila~\cite{hdvila}, and MiraData~\cite{miradata}, encompassing a wide range of video types: movies, egocentric videos, news, interviews, and how-to videos, and other in-the-wild videos of long duration. For data curation, we generate dense event labels for each long video. Specifically, we utilize existing high-quality short video captions (Panda-70M~\cite{panda70m} for HD-VILA~\cite{hdvila}, CosMo~\cite{cosmo} for HowTo100M~\cite{howto100m}, Ego4D-HCap~\cite{videorecap} for Ego4D~\cite{ego4d}, and the original high-quality captions provided in MiraData~\cite{miradata}) and filter the consecutive segments that can be regroup into a long video sequence, then we construct a sequence of event labels with their corresponding timestamps for every long video based on their captions. In this process, for datasets with high-quality event-level annotations (HT-Step~\cite{htstep} for HowTo100M~\cite{howto100m}, Ego4D-HCap~\cite{videorecap} for Ego4D~\cite{ego4d}), we directly utilize them as the event labels, while for others, we extract the major event from the caption using an LLM. Finally, we construct several types of long video QA pairs based on the video captions, event labels, and the timestamps of short video segments.They are categorized into five tasks: video captioning, temporal grounding, event relation recognition, scene relation recognition, and video event counting. See Appendix for more details.

\subsection{Multi-stage Short-to-Long Learning}
Unlike studies \cite{longva,longvila} that use long-form text to extend the context window, we prefer that direct training on long-form videos minimizes the gap between training and testing, leading to better downstream evaluations. Using a short-to-long scheme, our proposed \sysname{} is trained on a mixed dataset of both short and long videos. The training data are detailed in the Appendix.

\paragraph{Stage-1: Video-Language Alignment.} In this stage, we freeze the visual encoder and the large language model while training the compressor and the MLP to align the language with the compressed visual features. We use 0.5 million image-text pairs and 0.5 million short video-text pairs, and sample 4 frames from each video in training.

\paragraph{Stage-2: Short Video Pre-training.} To enhance the model's understanding of visual concepts, we conduct visual pre-training using 3.5 million images and 2.5 million short video-text pairs. Note most captions are refined using AI models to ensure richer and more detailed descriptions. During this stage, we sample 8 frames from each video.
 
\paragraph{Stage-3: Joint Short \& Long Video Instruction Tuning.} To enable the model to handle a wide variety of video tasks, we collect 3.5 million instruction fine-tuning samples, including 1.1M images, 1.7M short videos (under 60 seconds), and 0.7M long videos (60$\sim$3600 seconds). We mix the short and long video data to ensure the model retains fine-grained understanding while expanding its comprehension of long videos.
The sampling method used is the duration-based sampling described in Section~\ref{sec3_1}, with the number of sampled video frames ranging from 64 to 512.

\paragraph{Stage-4: Efficient High-Resolution Post-finetuning.} To enable the model to perceive higher resolutions, we employ a highly efficient post-finetuning strategy to adapt the original low-resolution video encoder to higher-resolution inputs. Specifically, we increase the input resolution of the video encoder from 224 to 448, freeze the LLM, and directly utilize 25\% of the stage-3 data for post-finetuning the video encoding. We find that this simple, full-data strategy effectively enhances the video encoder's adaptability to higher-resolution video inputs.

\subsection{Multi-Hop Needle in A Video Haystack} \label{sec_mhop}

\begin{figure}[!t] 
	\centering
	\includegraphics[width=1\linewidth]{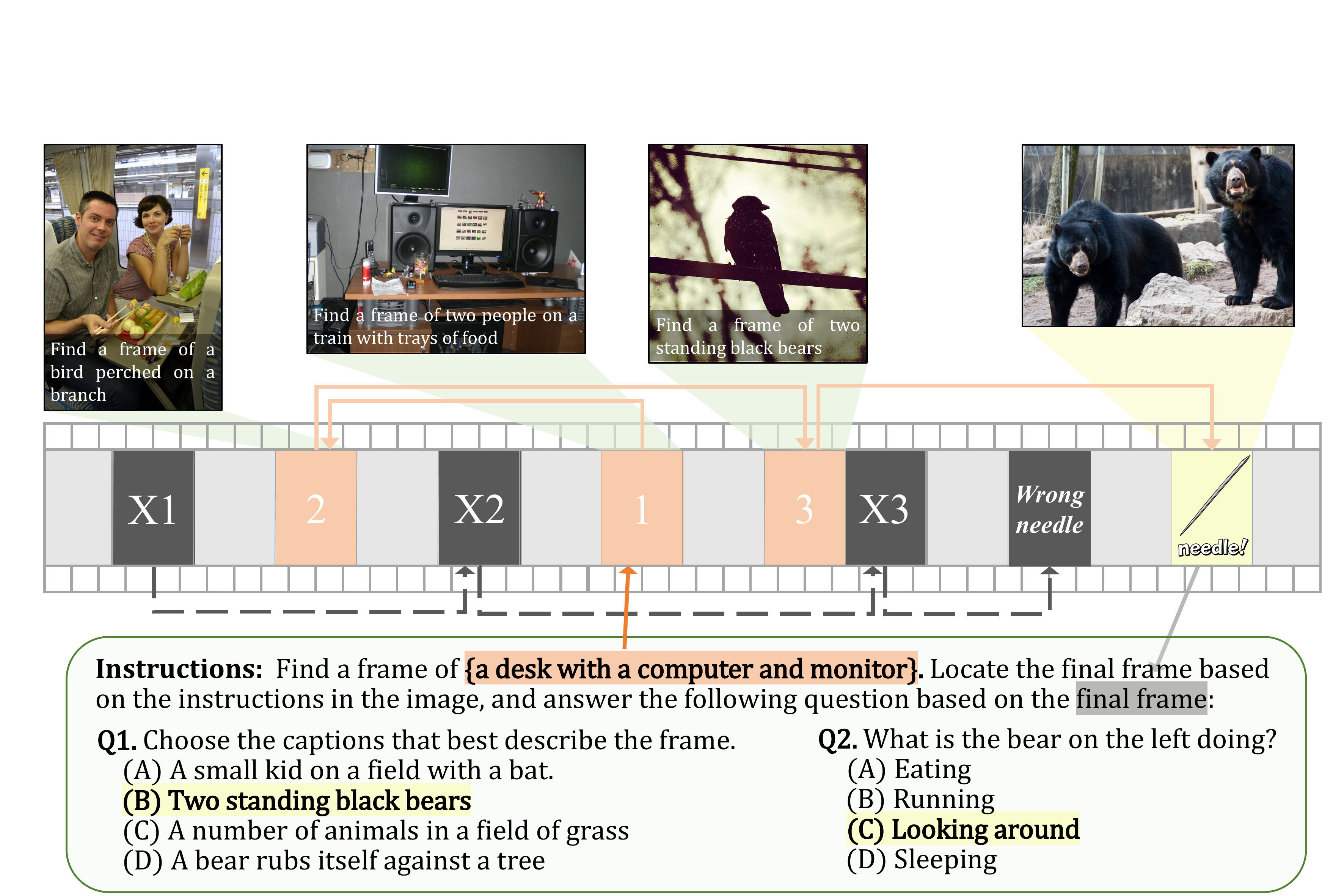}
		\caption{\textbf{An example of our Multi-Hop Needle in a Video Haystack}. The right path \textbf{\scalebox{0.75}{\colorbox{rightpath}{\textcolor{white}{(1, 2, 3)}}}} is for finding the needle while the wrong path \textbf{\scalebox{0.75}{\colorbox{wrongpath}{\textcolor{white}{(X1, X2, X3)}}}} is for distraction. MLLMs are asked to both find the needle (Q1) and answer its related question (Q2).}
    \label{niah_method}
\end{figure} 

Previous works~\cite{longva,longvila} utilize the ``Needle in a Video Haystack" (NIAH-Video) to evaluate the long video context understanding ability of models. Specifically, an image (commonly referred to as the ``needle") was inserted into a long video, and the model under test was then required to input the entire video and answer questions related to the needle. NIAH-Video assesses the model's capability to retrieve information from long videos. However, it has several drawbacks. Firstly, it is difficult to prevent images and questions similar to the needle from appearing in the model's training data, which leads to information leakage. Secondly, merely examining the model's visual retrieval ability is insufficient and lacks discrimination for evaluating its long video context understanding ability (many models can achieve an accuracy rate over 99\%). There is a need to further evaluate its reasoning ability regarding the content.

To address the above issues, we have designed a new evaluation task called ``Multi-Hop Needle in a Haystack" (MH-NIAH-Video). As shown in \cref{niah_method}, we insert a reasoning path composed of multiple images into the video haystack. Each image in this path has a randomly insertion position and corresponding textual clues to help find the next image. Given the starting point of the reasoning path, the model needs to follow this path to find the needle and answer questions related to it. What's more, to prevent the model from skipping the step of finding the needle by relying on information leakage or memorizing the content of all images, we insert multiple wrong reasoning paths simultaneously while inserting the correct reasoning path. The model needs to find the correct needle (Q1) along the correct reasoning path based on the given starting point and then answer questions related to the needle (Q2). In a way, our multi-hop approach offers a much more robust evaluation of the long context understanding ability in Multimodal Large Language Models (MLLMs) compared to the previous NIAH-Video. In practice, all images are sourced from MS-COCO~\cite{mscoco}, making use of its human-annotated captions and question-answer pairs. It should be noted that even if the model can perfectly remember the content of MS-COCO, it will not be of much help in finding the needle, which significantly reduces the likelihood of successful ``cheating".

\section{Experiments}
\begin{table*}[!htbp]
    \centering
\begin{adjustbox}{width=\linewidth,center}
\renewcommand{\arraystretch}{1.1}
\setlength{\tabcolsep}{1.5mm}
\begin{tabular}{lrrcccccccccccc}
\toprule  \multirow{2}{*}{\textbf{Model}} & \multicolumn{1}{c}{\multirow{2}{*}{\centering \textbf{Size}}} & \multicolumn{1}{c}{\textbf{Avg tokens}}  & {\textbf{MVBench}} & {\textbf{PerceptionTest } }  &{\textbf{LongVideoBench}} & {\textbf{MLVU}} & \multicolumn{2}{p{4cm}}{\centering \textbf{VideoMME (\textit{w/o \& w sub.})} } & {\textbf{LVBench}}  & {\textbf{Charades-STA}}& {\textbf{AuroraCap}}\\ \cline{8-9}
&&  \multicolumn{1}{c}{\textbf{per frame}} & Avg &Val&Val&M-Avg& Overall & Long & Avg & mIoU & Avg  \\
\rowcolor{gray!10} Avg. Duration & &  & 16s& 23s  & 473s & 651s &  1010s & 2386s & 4101s & 30s & 28s \\
\midrule
\textit{Proprietary Models} & & \\
GPT-4V~\citep{gpt4v} & - & - & 43.7 & - & 59.1  & 49.2 & 59.9/63.3 & 53.5/56.9 & - & - & - \\
GPT-4o~\citep{gpt4o} & - & - & 64.6 & - & 66.7  & 64.6 & 71.9/77.2 & 65.3/72.1 & 30.8 & 35.7 & - \\
Gemini-1.5-Pro~\citep{gemini} & - & - & 60.5 & - & 64.0  & - & 75.0/81.3 & 67.4/77.4  & 33.1 & - & 41.7\\
\midrule
\textit{Small Size MLLMs} & & \\
Qwen2-VL~\citep{qwen2vl}& 2B  & 1924 & 63.2  & -  &  -  & -  & 55.6/60.4 & - & - & - & - \\
InternVL2.5~\citep{internvl2_5}& 2B & 256 & 68.8  & -  &  46.0  & 61.4  & 51.9/54.1 & -  & - & - & - \\
\rowcolor{darkGreen!20} \textbf{\modelname~}@448 &2B & 16 & \textbf{70.0} & \textbf{70.5} & \textbf{58.3}   & \textbf{65.7} & \textbf{57.0/63.9} & \textbf{44.9/54.0} & \textbf{42.9} & \textbf{45.2} & - \\
\midrule
\textit{Open-Source MLLMs} \\

VideoChat2-HD~\citep{videochat2} & 7B & 72 & 62.3 & - & -  & 47.9 & 45.3/55.7 & 39.8/53.9 & - & 3.4 & - \\
InternVideo2-HD~\citep{internvideo2} & 7B & 72 & 67.2 & 63.4 & -  & - & 49.4/\space\space\space-\space\space\space & - & - & - & - \\

LLaVA-OneVision~\citep{llavaonevision} & 7B & 196 &56.7 & 57.1 & 56.3  & 64.7 & 58.2/61.5 & - & - & 13.5 & 37.5 \\
\color{gray} LLaVA-OneVision~\citep{llavaonevision} & \color{gray} 72B & \color{gray}196 &\color{gray}59.4 & \color{gray}66.9 & \color{gray}61.3  & \color{gray}68.0 & \color{gray}66.2/69.5 & - & - & - & - \\
LLaVA-Video~\citep{llavavideo} & 7B & 676 & 58.6 & 67.9 & 58.2  & 70.8 & 63.3/69.7 & - & - & - & 39.0 \\
VITA1.5~\cite{vita} & 7B & 256 & 56.8 & - & -  & - & 56.8/59.5 & - & - & - & - & - \\
InternVL2~\citep{internvl2} & 8B & 256 & 65.8 & - & 54.6  & 64.0 & 54.0/56.9 & - & - & - & 37.7 \\
\color{gray}InternVL2~\citep{internvl2} & \color{gray}76B & \color{gray}256 & \color{gray}69.6 & \color{gray}- & \color{gray}61.1  & \color{gray}69.9 & \color{gray}61.2/62.8 & \color{gray}- & - & - & - \\
InternVL2.5~\citep{internvl2_5} & 8B & 256 & 72.0 & - & 60.0  & 68.9 & 64.2/66.9 & - & -  & - & - \\
Qwen2-VL~\citep{qwen2vl} & 7B & 1924 & 67.0 & 66.9 & -  & - & 63.3/69.0 & - & -  & - & 41.6 \\
Qwen2.5-VL~\citep{qwen25vl} & 7B & 1924 & 69.6 & - & 56.0  & 70.2 & 65.1/71.6 & - & 45.3  & 43.6 & - \\

\midrule
\textit{Open-Source Long Video MLLMs}  \\

LLaMA-VID~\citep{llamavid} & 7B & 2 & 41.9 & 44.6 & - & 33.2 & 25.9/\space\space\space-\space\space\space & - & 23.9 & - & 30.9 \\

LongVU~\citep{longvu} & 7B & 64 & 66.9 & - & -  & 65.4 & \space\space\space-\space\space\space/60.6& \space\space\space-\space\space\space/59.5 & - & - & - \\
LongVA~\citep{longva} & 7B & 144 & - & - & -  & 56.3 & 52.6/54.3 & 46.2/47.6 & - & - & 34.5 \\
LongVILA~\citep{longvila} & 7B & 196 & 67.1 & 58.1 & 57.1  & - & 60.1/65.6 & 47.0/52.1 & - & - & - \\
Kangaroo~\cite{kangaroo} & 8B & 256 & 61.0 & - & 54.8 & 61.0 & 56.0 / 57.6 & 46.7 / 49.3 & 39.4 & - & - \\
\rowcolor{darkGreen!20} \textbf{\modelname~}@224& 7B & 16 & 73.2 & 75.6 & 64.2  & 74.5 & 64.0/69.4 & 53.6/61.9 & 47.2 & \textbf{48.4} & - \\
\rowcolor{darkGreen!20} \textbf{\modelname~}@448 & 7B & 16 & \textbf{74.0} & \textbf{76.2} & \textbf{64.7}   & \textbf{74.7} & \textbf{65.3/69.7} & \textbf{55.4/63.3} & \textbf{48.2} & 48.0 & \textbf{42.9}\\

\bottomrule
\end{tabular}
\end{adjustbox}
\caption{\textbf{Results on comprehensive video-linguistic benchmarks}}
\label{tab:main}
\vspace{-2mm}
\end{table*}
\paragraph{Implementation details.} 
We employ UMT-L~\cite{umt}, token merging with MLP, and Qwen2-7B as visual encoder, connector, and LLM, respectively. When processing a long video, we divide it into shorter clips, each consisting of 4 frames. Each clip is compressed into 64 tokens, meaning that, on average, each frame is represented by 16 tokens. Regarding video-level compression, while it presents some challenges in compatibility with training acceleration strategies such as sequence parallelism, we only employ it during inference. In most of the ablations, we use only one-fourth of the full dataset. See Appendix for details.

\subsection{General Video Understanding Evaluation}

\paragraph{Benchmark.} We evaluate our model on six general video understanding benchmarks in question-answering format, including two short video benchmarks: MVBench~\cite{videochat2} and Perception Test~\cite{perceptiontest}, and three long video benchmarks: LongVideoBench~\cite{longvideobench}, MLVU~\cite{mlvu} and LVBench~\cite{lvbench}, and a comprehensive benchmark, VideoMME~\cite{videomme}, covering videos ranging from minute-level to hour-level durations. We further evaluate the temporal grounding and video caption tasks, using the Charades-STA~\cite{charades_sta} and AuroraCap~\cite{videorecap}.

\paragraph{Leading performance.} As depicted in \cref{tab:main}, our \modelname{} achieves the best results on diverse VideoQA benchmarks within the 2B and 7B size category, significantly outperforming other approaches. Remarkably, its performance even eclipses that of models with substantially larger scales, such as InternVL2-76B, as well as proprietary models like GPT-4o and Gemini-1.5-Pro. Even when merely supplying timestamp information via a text prompt, our model has achieved remarkable performance in temporal grounding. Meanwhile, it also significantly outperforms other models in the video captioning task, even surpassing the proprietary GPT-4o and Gemini-1.5 Pro. This demonstrates the effectiveness of the comprehensive design of our model, data, and training strategies.

\begin{figure}[!t] 
	\centering
    	\begin{subfigure}{1.\linewidth}
		\centering
		\includegraphics[width=1\linewidth]{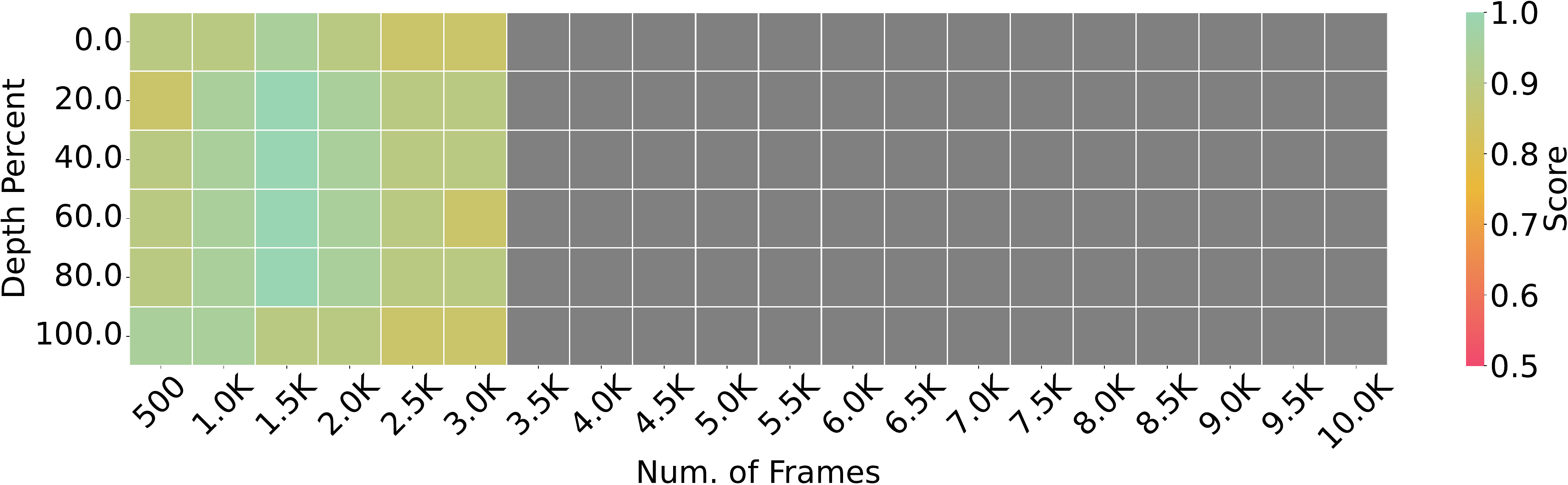}
		\caption{LongVA~\cite{longva}, accuracy=91.8\% at 3k frames}
	\end{subfigure}
    
	\begin{subfigure}{1.\linewidth}
		\centering
		\includegraphics[width=1\linewidth]{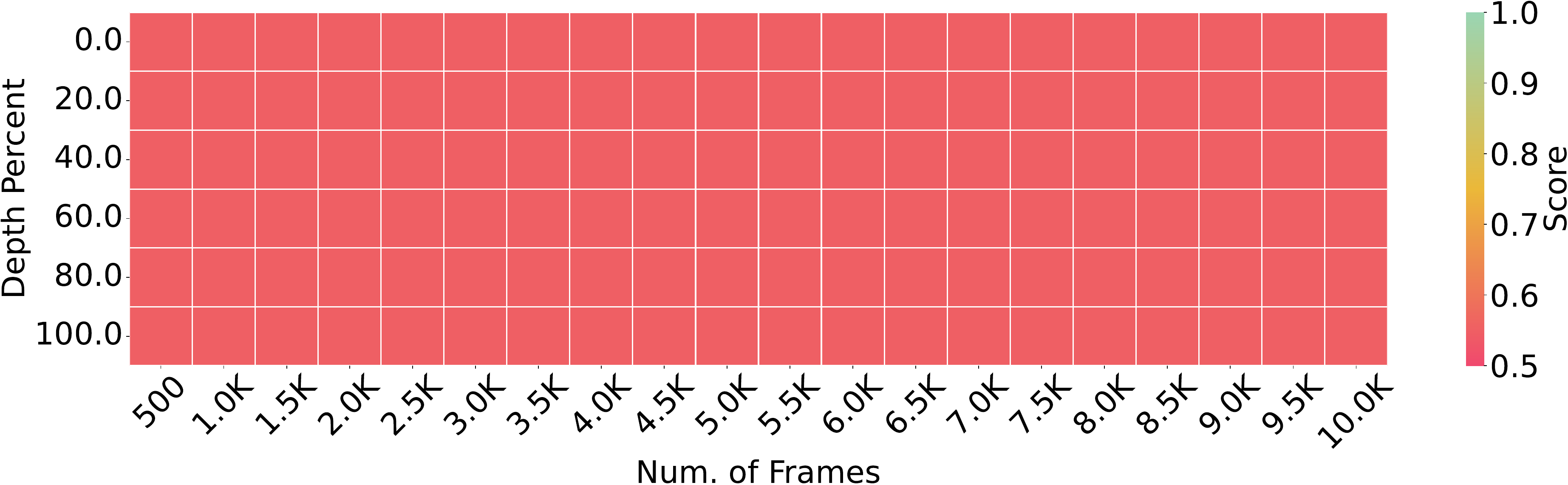}
		\caption{LLama-VID~\cite{llamavid}, accuracy=55.0\% at 10k frames}
	\end{subfigure}
	
	\begin{subfigure}{1.\linewidth}
		\centering
		\includegraphics[width=1\linewidth]{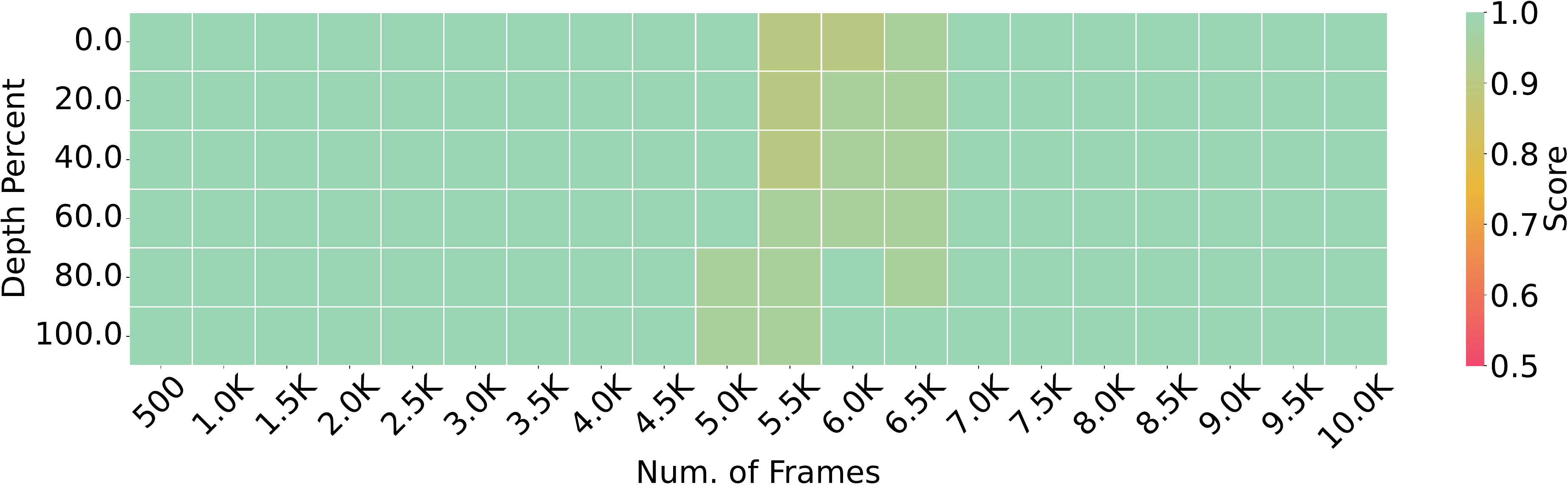}
		\caption{\modelname{} (ours), accuracy=\textbf{99.1}\% at 10k frames}
	\end{subfigure}

    \caption{\textbf{Results on the ``Single-Hop Needle-in-A-Video-Haystack" evaluation with 10,000 frames.}}
    \label{niah_single}
    \vspace{-4mm}
\end{figure}

\subsection{Long Video Context Evaluation}
\paragraph{Baseline.} LongVA~\cite{longva} and LLama-VID~\cite{llamavid} are used as baselines. LongVA trains MLLMs using long text data, transfering the long context of LLM from text to video. LLama-VID accomplishes efficient inference of long videos by compressing each frame to only two tokens. Our model benefits from these two, so we take them as baselines.

\paragraph{Single-Hop NIAH.} As shown in Fig. \ref{niah_single}, we follow the protocols in LongVA \cite{longva} for Single-Hop NIAH, we source a long video and sample frames uniformly from it. Then we add needles (indicating images) into the sampled image sequence at different positions. MLLMs are fed with this long image sequence and answer the corresponding questions to the indicating images. We evaluate all models over 10,000 frames. Note our \modelname{} delivers a 99.1\% success rate in accurately retrieving the correct indicating image and answering the related question even across 10,000 frames. In comparison, LongVA gives a decent result close to 92\% within 3000 frames while LLama-VID only achieves 55\% accuracy. It demonstrates \modelname{}'s state-of-the-art performance in long multimodal context modeling.

\begin{figure}[!t] 
	\centering
	\includegraphics[width=1.\linewidth]{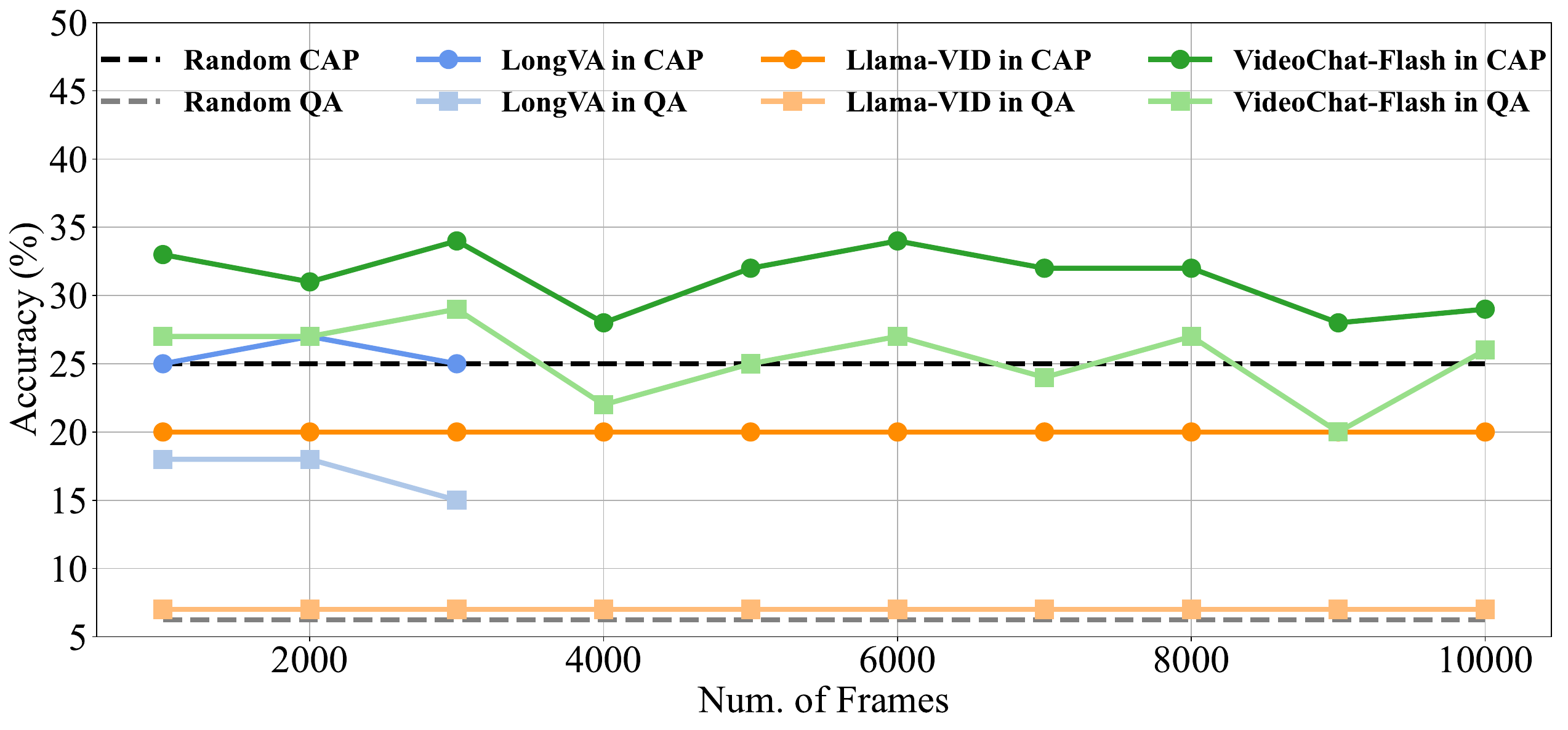}
    \caption{\textbf{Results on the ``Multi-Hop Needle-in-A-Video-Haystack with 10,000 frames.}}
    \label{niah_multiple}
    \vspace{-4mm}
\end{figure} 

\paragraph{Multi-Hop NIAH.} In this evaluations, MLLMs need to trace along the chain of indicating images, locate the needle, and answer its question. Two metrics ``CAP" and ``QA'' are used to denote the accuracy of finding the correct needle and the accuracy of answering the questions related to the needle as well as finding the needle successfully, respectively. As shown in \cref{niah_multiple}, our \modelname{} still beats all baselines. Specifically, \modelname{} gives 31.3\% and 25.4\% in ``CAP'' and ``QA'' on average, higher than LongVA by around 8 points. It can be seen that compared with the single-hop NIAH, the multi-Hop NIAH presents a much more difficult challenge, which can better reflect the real gap between the capabilities of different models.

\subsection{Ablation \& Analysis}\label{sec:ablation}

\paragraph{Effect of various designs.} As shown in \cref{tab:ablation_training}, we have conducted comprehensive ablation studies on each design. In terms of the model, it can be observed that HiCo significantly reduces the computational load (from 196 to 16 tokens per frame) while barely compromising the performance. Meanwhile, duration-based sampling and timestamp prompts play crucial roles in enhancing the performance. The further leap in performance mainly stems from the training strategy in short-to-long learning and a better mixture of training data.

\begin{table}[t]
    \centering
    \resizebox{1.0\linewidth}{!}{
        \begin{tabular}{l|rrrr}

        \multirow{2}{*}{\textbf{Settings}} & \textbf{\small{MVB}} & \textbf{\small{MLVU}} & \textbf{\small{VMME}} & \textbf{\small{Charades}} \\
         & \small{Avg} & \small{\small{M-Avg}} & \small{Overall} & \small{mIoU} \\
        
             \toprule
        Baseline & 60.2 & 63.7 & 52.8 & \\
        \rowcolor{darkGreen!20}+ HiCo & 61.1 & 60.6 & 53.2 & - \\
        \rowcolor{orange!20}+ short video pretraining & 66.5 & 62.4 & 53.9 & - \\
        \rowcolor{darkGreen!20}+ duration-based sampling & 67.0 & 64.5 & 55.5 & - \\
        \rowcolor{orange!20}+ LongVid data & 66.5 & 68.3 & 55.8 & - \\
       \rowcolor{orange!20}  + Joint short \& long sft & 73.2 & 74.5 & 64.0 & 48.4 \\
       \rowcolor{red!20} + High-res post ft & 74.0 & 74.7 & 65.3 & 48.0 \\
       \rowcolor{darkGreen!20} - timestamp prompt & 73.4 & 73.2 & 63.4 & 44.2 \\
        \end{tabular}
    }
    \caption{\textbf{Effect of various designs} on \scalebox{0.85}{\colorbox{orange!20}{data}}, \scalebox{0.85}{\colorbox{darkGreen!20}{model}}, and \scalebox{0.85}{\colorbox{red!20}{resolution}}. The baseline employs SigLiP-so400M~\cite{siglip} as the vision encoder and Spatial donwsampling (196 tokens per frame) as the connector. It adopts a two-stage training strateay with image and short video following LLaVA~\cite{llava}.}
    
    \label{tab:ablation_training}
    \vspace{-4mm}
\end{table}

\begin{figure}[ht]
    \begin{subfigure}[b]{0.44\textwidth}
        \centering
        \includegraphics[width=1.\linewidth]{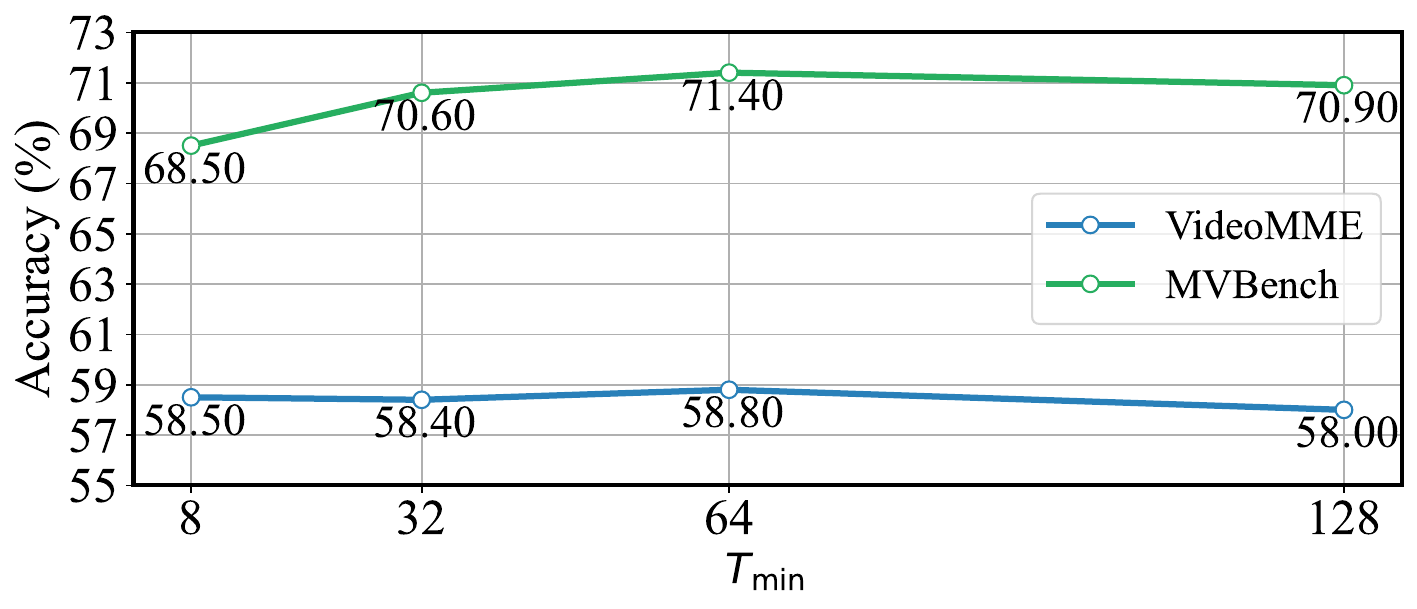}
        \caption{Effect of $T_{\text{min}}$ ($T_{\text{max}}$=128)}
    \end{subfigure}
    \begin{subfigure}[b]{0.44\textwidth}
        \centering
        \includegraphics[width=1.\linewidth]{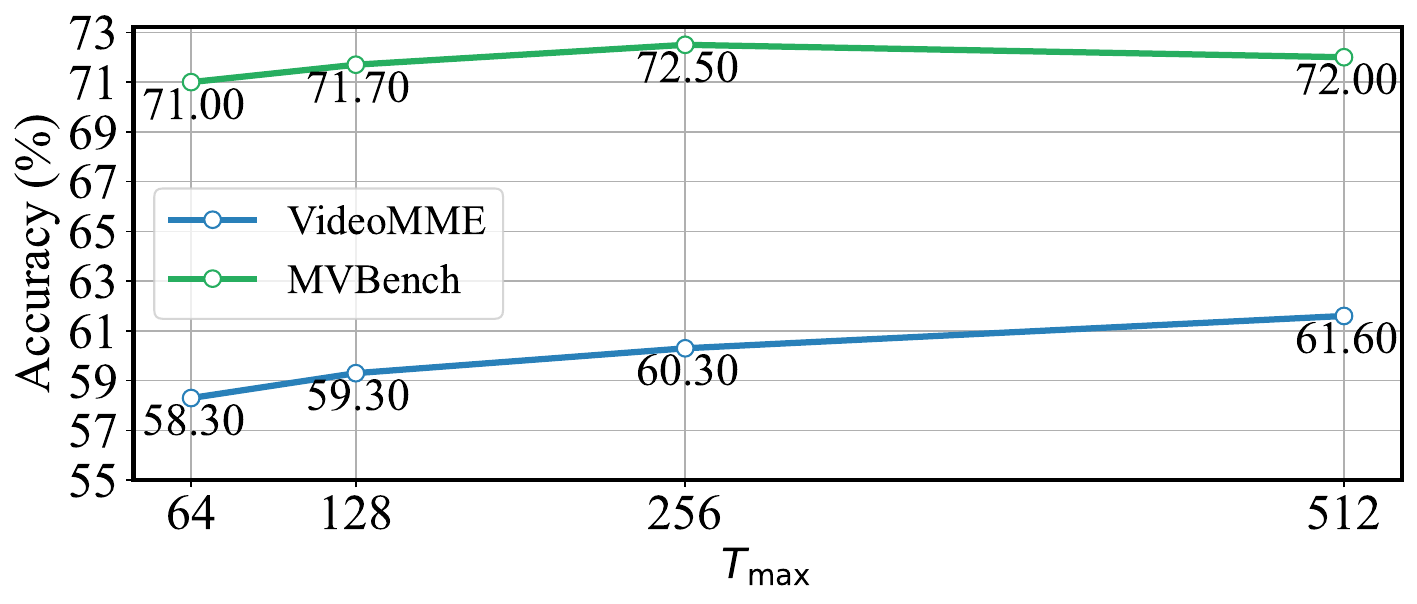}
        \caption{Effect of $T_{\text{max}}$ ($T_{\text{min}}$=64)}
    \end{subfigure}
    \caption{\textbf{Ablation of Duration-based Sampling}}
    \label{fig:sampling}
    \vspace{-5mm}
\end{figure}

\paragraph{Duration-based Sampling.} As shown in \cref{fig:sampling}, A relatively large $T_{\text{min}}$ (64) enables the model to better learn to model the fine actions and rapid movements in short videos during training, thereby enhancing the performance of short video understanding. Increasing $T_{\text{max}}$ from 64 to 256 leads to a stable improvement in the understanding performance of both short and long videos. This indicates that more sampled frames can extract more accurate information from our long video data. When $T_{\text{max}}$ reaches 512, there is a slight decline in the performance of short videos. Overall, it achieves a balance between the performance of short and long videos.

\paragraph{Video encoders are efficient clip compressors.} As shown in \cref{tab:ablation_vit}, we have tested the most popular image encoder, SigLIP~\cite{siglip}, and the short video encoder, UMT~\cite{umt}, for encoding clips with heavy compression. We found that even when the computational cost is significantly lower, UMT can still achieve better performance on the short video task MVBench. Moreover, as the size of the training data increases from 2 million to 8 million, UMT outperforms SigLIP distinctly across various benchmarks. We believe that this is attributed to the spatio-temporal attention employed by UMT, which can aggregate the key information from different frames within a clip, thus enabling the learning of more compact compression features.

\begin{table}
    \centering
    \setlength\tabcolsep{1.0pt}
    \resizebox{1\linewidth}{!}{

         \begin{tabular}{l|cc|lll}

\multirow{2}{*}{\textbf{Visual Encoder}}& \textbf{FLOPs} & \textbf{Latency}  & {\textbf{MVBench}} & {\textbf{MLVU}} & {\textbf{VideoMME}} \\

~ & (G) & (ms)  & Avg & M-Avg & Overall \\

\toprule

\multicolumn{6}{l}{\darkGreen{\textit{\#tokens per frame=16,  training data size=2M}}} \\
SigLIP$_\mathrm{SO400M}$@384 & 2679 & 79.7 & 60.2  & 62.0 & \textbf{53.5} \\
UMT-L@224 & 596 & 11.8 & 61.1\textcolor{red}{(+0.9)} & 60.0\textcolor{teal}{(-2.0)} & 53.2\textcolor{teal}{(-0.3)} \\
\midrule
\multicolumn{6}{l}{\darkGreen{\textit{\#tokens per frame=16,  training data size=8M}}} \\
SigLIP$_\mathrm{SO400M}$@384 & 2679 & 79.7 & 71.2  & 70.8 & 62.4 \\
UMT-L@224 & 596 & 11.8 & 73.5\textcolor{red}{(+2.3)} & 73.7\textcolor{red}{(+2.9)} & 62.7\textcolor{red}{(+0.3)} \\

\end{tabular}
}

    \caption{\textbf{Comparison of visual encoders.
    }
    }
    \label{tab:ablation_vit}
    \vspace{-1mm}
\end{table}

\paragraph{Different connectors and compression ratio.} As shown in the \cref{tab:ablation_compressor}, we consider three different numbers of tokens per frame after compression (16, 49, 196) and four popular token compression strategies: spatial downsampling~\cite{llavanextvideo,internvl2}, uneven downsampling~\cite{uneven}, spatio-temporal resampler~\cite{videochat2,koala}, and similar token merging~\cite{tome,longvlm} (more details can be found in the Appendix). It can be seen that compared with other methods, the parameter-free similar token merging operation can achieve a remarkably low compression ratio and even obtain better performance than without compression. Even in the extreme case of a 2\% compression ratio, it can still maintain most of the performance.

\begin{table}
    \centering
    \setlength\tabcolsep{3.0pt}
    \resizebox{1\linewidth}{!}{
 
        \begin{tabular}{l|ccc|l}

\multirow{2}{*}{\textbf{Connector}}& \multirow{1}{*}{\textbf{MVBench}} & \multirow{1}{*}{\textbf{MLVU} }& \multirow{1}{*}{\textbf{VideoMME}} & \multirow{2}{*}{\textbf{Avg}}\\

~ & Avg & M-Avg & Overall & ~ \\

\toprule

\multicolumn{5}{l}{\darkGreen{\textit{\#tokens per frame=729, compression ratio=100\%}}} \\
MLP (Uncompressed) & 59.4 & 64 & 55.3 & 59.6 \\

\midrule
\multicolumn{5}{l}{\darkGreen{\textit{\#tokens per frame=196, compression ratio=27\%}}} \\
Spatial Downsampling & 60.2 & 63.7 & 52.8  & 58.9(-0.7)\\
Uneven Downsampling& 60.9  & 62.5 & 54.9 & 59.4(-0.2)\\
Spatio-temporal Resampler & 59.5 & 61.9 & 51.9 & 57.8(-1.8)\\
\rowcolor{gray!20} 
\textbf{Similar Token Merging} & \textbf{62.8}  & \textbf{66.7} & \textbf{56.8}  & \textbf{62.1}\textcolor{red}{(+2.5)} \\

\midrule
\multicolumn{5}{l}{\darkGreen{\textit{\#tokens per frame=49, compression ratio=7\%}}} \\
Spatial Downsampling & 60.2 & 61.8 & 53.6  & 58.5(-1.1)\\
Uneven Downsampling& 59.8  & 62.8 & 54.3 & 59.0(-0.6) \\
Spatio-temporal Resampler & 55.5 & 58.1 & 51.1 & 54.9(-4.7) \\
\rowcolor{gray!20} 
\textbf{Similar Token Merging} & \textbf{61.4}  & \textbf{63.3} & \textbf{55.3}  & \textbf{60.0}\textcolor{red}{(+0.4)}  \\

\midrule
\multicolumn{5}{l}{\darkGreen{\textit{\#tokens per frame=16, compression ratio=2\%}}} \\
Spatial Downsampling & 58.1 & 61.1 & 50.1  & 56.4(-3.2) \\
Uneven Downsampling& 58.3 & 60.0 & 52.3 & 56.9(-2.7) \\
Spatio-temporal Resampler & 51.4 & 54.7 & 47.7 & 51.3(-8.3) \\
\rowcolor{gray!20} 
\textbf{Similar Token Merging} & \textbf{60.2}  & \textbf{62.4} & \textbf{53.5}  & \textbf{58.7}\textcolor{teal}{(-0.9)} \\

\end{tabular}
}

    \caption{\textbf{Comparison of connectors.}
    }
    \label{tab:ablation_compressor}
       \vspace{-4mm}
\end{table}

\paragraph{Progressive visual dropout.}  As shown in the \cref{tab:ablation_llm}, at the shallow layers of the LLM, uniform dropout performs better than attention select on long video tasks. However, at the deep layers of the LLM, attention select shows better performance. Performing visual dropout at the deep layers can not only improve the computational efficiency but also enhance the performance. Combining uniform dropout and attention select can achieve a good balance between performance and efficiency. More relevant analyses and comparative experiments can be found in the Appendix.

\begin{table}
    \centering
    \setlength\tabcolsep{2.0pt}
    \resizebox{1\linewidth}{!}{

\begin{tabular}{l|c|cc|ll}

\multirow{2}{*}{\textbf{Drop type/keep ratio}}&  \multirow{2}{*}{\textbf{Drop layer}}&  \multicolumn{1}{c}{\textbf{FLOPs}} & \multicolumn{1}{c|}{\textbf{Latency}} & {\textbf{MLVU}} & {\textbf{VideoMME}}\\

~ &  & \multicolumn{1}{c}{(G)} & \multicolumn{1}{c|}{(s)} & M-Avg & Overall \\
\toprule
- & - & 341.4 & 2.6  & 71.8 & 61.2 \\
\midrule
Uni./0.5 & 4 & 242.8 & 1.9 & \textbf{71.2} & 60.4 \\
Attn./0.5 & 4 & 242.8 & 1.9 & 70.7 & \textbf{60.8} \\
\midrule
Uni./0.5 & 18 & 295.2 & 2.2 & 71.7 & \textbf{61.8} \\
\rowcolor{gray!20} 
Attn./0.5 & 18 & 295.2 & 2.2 & \textbf{72.1}\textcolor{red}{(+0.3)} & 61.7\textcolor{red}{(+0.5)} \\
\midrule
Attn./0.75,Attn./0.25 & 4,18 & 245.8  & 1.9 & 71.4  & 60.9  \\
\rowcolor{gray!20} 
Uni./0.75,Attn./0.25 & 4,18 &  245.8 & 1.9 & \textbf{72.0}\textcolor{red}{(+0.2)}   & \textbf{61.1}\textcolor{teal}{(-0.1)} \\

\end{tabular}
}

    \caption{\textbf{Effectiveness  of visual dropout.} The Qwen2-7B we used has a total of 28 layers. "Uni." and "Attn." represent uniform drop and attention select respectively.}
    \label{tab:ablation_llm}

\end{table}

\vspace{-0.3cm}

\begin{table}
    \centering
    \setlength\tabcolsep{2.0pt}
    \resizebox{\linewidth}{!}{     \begin{tabular}{l|l|rr|rr}
        \textbf{Input} & \multirow{2}{*}{\textbf{Model}}& \textbf{Avg tokens} & \multicolumn{1}{c|}{\textbf{FLOPs}} & \multicolumn{2}{c}{\centering \textbf{Memory(G)}} \\
        \textbf{frames} &  &  \textbf{per frame} &  \multicolumn{1}{c|}{(T)}   & Train & Infer  \\
        \toprule
       \multirow{3}{*}{{64}} & LongVILA~\citep{longvila} & 196  & 224.8 & 15.4  & 16.7  \\
       & LongVA~\citep{longva} & 144  & 155.9  & 12.3 &  16.3   \\
    \rowcolor{darkGreen!20} \cellcolor{white}&     \textbf{\modelname{}} & \textbf{16}  & \textbf{14.8} & \textbf{4.8} & \textbf{15.4} \\
        \midrule
       \multirow{3}{*}{{256}} & LongVILA~\citep{longvila} & 196 &  1467.5 & 50.1 & 21.0    \\
      &  LongVA~\citep{longva} & 144  & 930.4 & 37.8 & 19.5 \\
     \rowcolor{darkGreen!20}\cellcolor{white}  &  \textbf{{\modelname{}}}& \textbf{16} &  \textbf{63.0} &  \textbf{7.6} & \textbf{{15.7}}   \\
        \midrule
        \multirow{3}{*}{{1000}} &  LongVILA~\citep{longvila} & 196 & 14336.9 & \textit{oom} & 37.7   \\
        & LongVA~\citep{longva} & 144  & 8278.9  & \textit{oom} & 31.8  \\
   \rowcolor{darkGreen!20} \cellcolor{white}   &   \textbf{{\modelname{}}} & \textbf{16}  & \textbf{303.3} & \textbf{18.6} & \textbf{17.1}    \\
        \midrule
       \multirow{3}{*}{{10000}} &  LongVILA~\citep{longvila} & 196 & 1184250.0 & \textit{oom} & \textit{oom}  \\
       &  LongVA~\citep{longva} & 144  & 644632.0 & \textit{oom} & \textit{oom}   \\     
     \rowcolor{darkGreen!20}  \cellcolor{white} &   \textbf{{\modelname{}}} & \textbf{16}  & \textbf{9969.5} & \textit{oom}  & \textbf{33.6}   \\
    \end{tabular}
}

    \caption{\textbf{Comparison of FLOPs and Cuda memory.} The FLOPs and inference memory is estimated using one NVIDIA A100-80G GPU with one sample, and the training is estimated using 32 NVIDIA A100-80G GPUs with DeepSpeed ZeRO-3~\cite{deepspeed}. We assume that the visual features have been extracted and stored in advance, so we only consider the FLOPs and memory of the LLM.
    }
    \label{tab:speed}
    \vspace{-0.3cm}
\end{table}

\paragraph{Model efficiency.} As in \cref{tab:speed}, even when processing short videos, the compute load of our model is only one-tenth that of previous models. Meanwhile, as the number of input frames increases, the difference becomes more and more pronounced. Only our model can complete the inference on 10,000 frames on a single A100-80G. Concretely, \modelname{}'s compute load is two orders of magnitude lower than that of LongVILA~\cite{longvila} (9,969.5 vs. 1,184,250.0 TFLOPs).

\section{Conclusions}
In this paper, we address the challenge of long-context video modeling in MLLMs from the model architecture, training data, training strategy and evaluation benchmark. We design an efficient architecture for video MLLMs by introducing a hierarchical long video context compression method, which achieves an extreme compression ratio with nearly no performance loss. Regarding data and training, we propose a new long video training corpus and short-to-long learning strategy, which effectively enhances the model's understanding ability for videos of various lengths. Additionally, we developed a new and more challenging long video context evaluation benchmark. Our model demonstrated outstanding performance on various video understanding benchmarks, which validates the effectiveness of our proposed methods.

{
    \small
    \bibliographystyle{ieeenat_fullname}
    \bibliography{main}
}
\maketitlesupplementary

\section{More Results \& Discussions}

\subsection{Visual Dropout in LLM}

\begin{figure}
    \centering
    \includegraphics[width=0.9\linewidth]{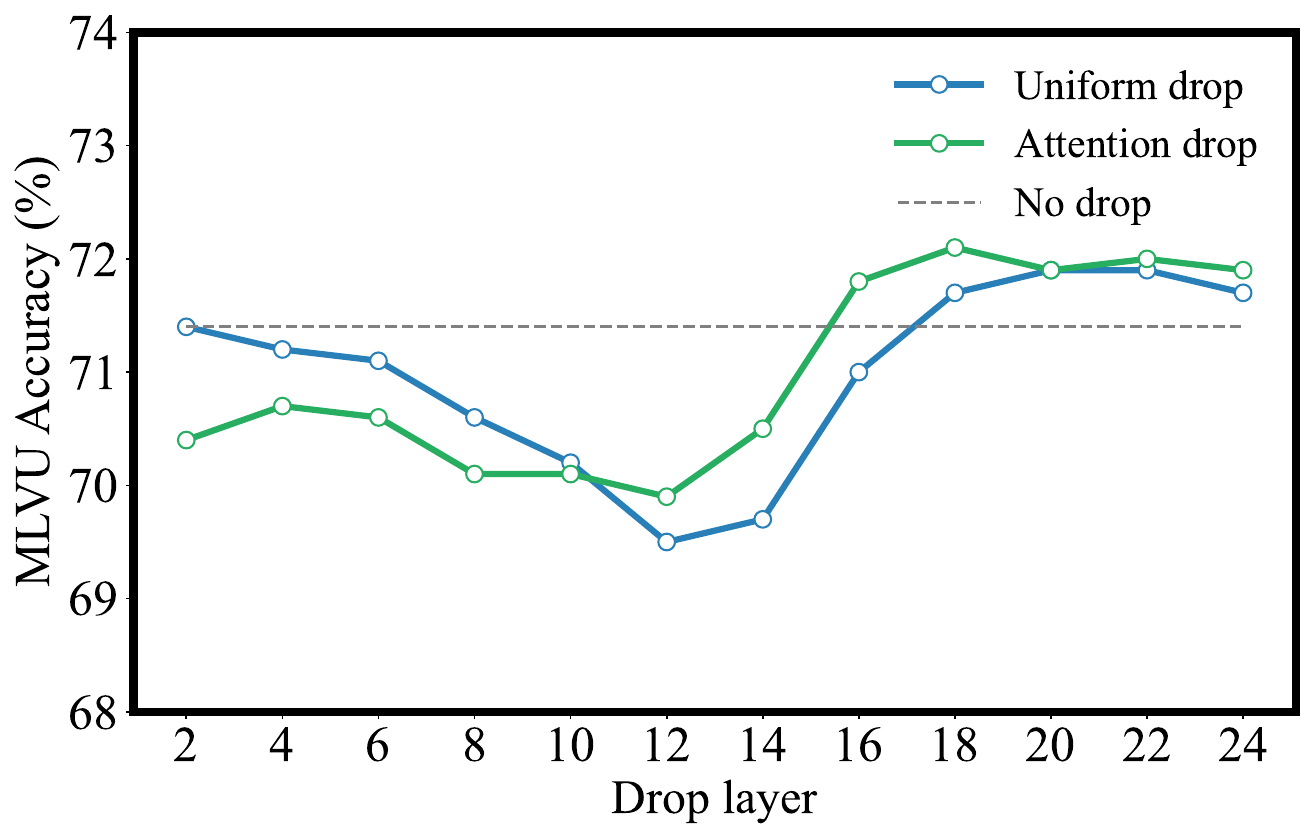}
    \caption{\textbf{Visual redundancy in long context across layers.} We conduct experiments on Qwen2-7B (28 layers) and test the impact of droping 50\% of the visual tokens from shallow to deep layers.}
    \label{fig:dropout}
\end{figure}
\paragraph{Visual token redundancy in LLM inference.} As shown in \cref{fig:dropout}, we find that even when half of the tokens are discarded at the shallow layers of the LLM, the performance of long video understanding only degrades marginally. This indicates that despite high compression at the clip level (encoding each frame into only 16 tokens), there remains considerable redundancies between clips when their representations are interacted in the LLM. Furthermore, we find the overall understanding performance gets better as the dropout happens in the deeper layer of the model. Remarkably, at approximately two-thirds of the LLM's depth, the performance even surpasses that of the no-discard baseline. This might suggest that in the deeper layers of the network, an excess of visual tokens may interfere with the model's reasoning process. For the drop type, we observe that uniform drop often outperforms attention-based selection in the shallow layers. We suppose, at these layers, the LLM has not yet fully determined the specific locations to focus on. As a result, relying on attention may introduce bias.

\begin{figure}
    \centering
    \includegraphics[width=1.0\linewidth]{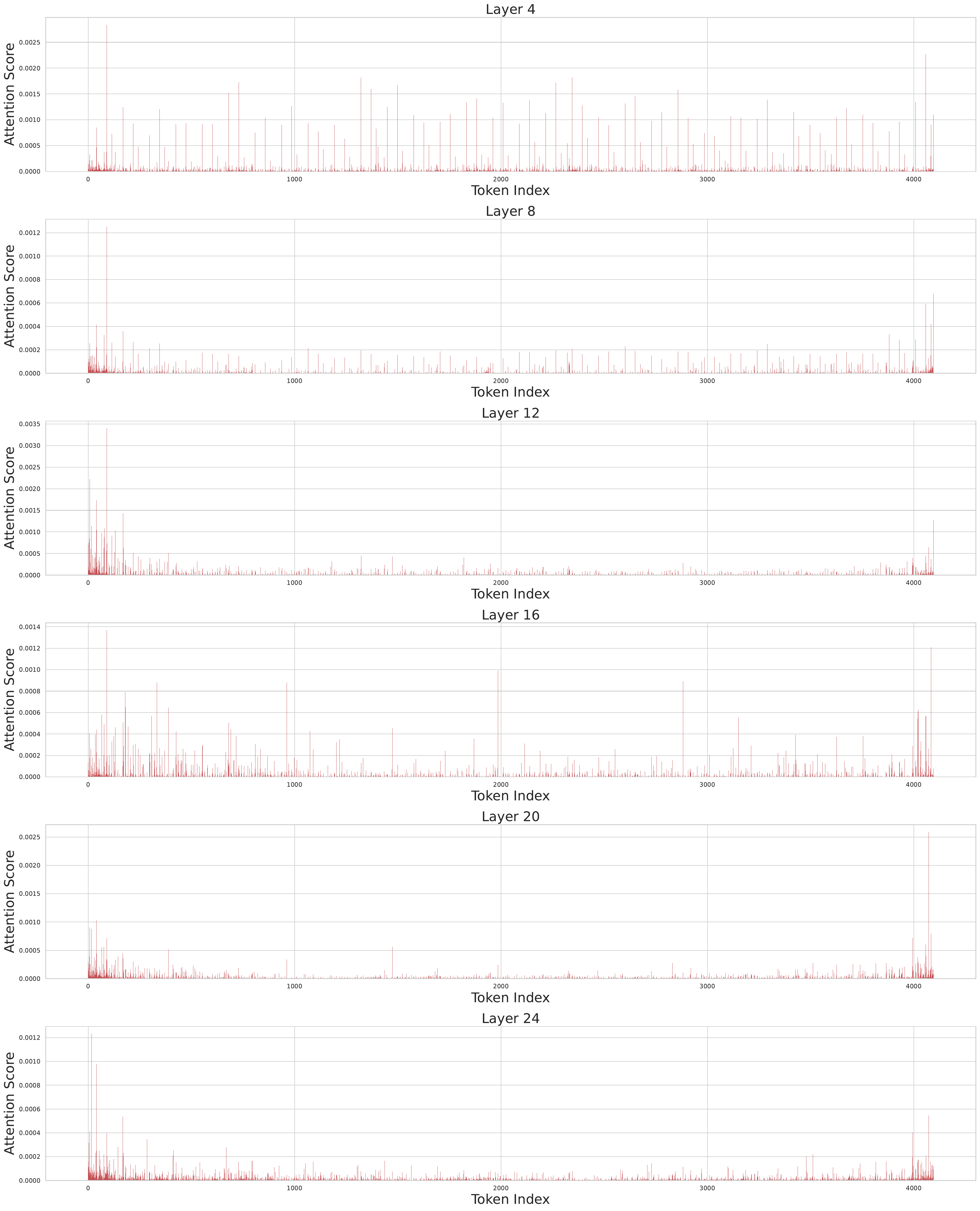}
    \caption{\textbf{Visualization of the attention scores from the last textual token to visual tokens at each layer of the network.}}
    \label{fig:attn_vis}
\end{figure}

\paragraph{Visualization of visual attention map.} As shown in the ~\cref{fig:attn_vis}, for long video context, the attention of text tokens is relatively dispersed in the shallow layers of the network. However, as the layers deepen, the attention gradually becomes focused on specific regions. Thus, we believe that the attention scores in the deeper layers are more reliable, while those in the shallow layers may be prone to bias.

\begin{table*}[!htbp]
    \centering
\begin{adjustbox}{width=\linewidth,center}
\renewcommand{\arraystretch}{1.1}
\setlength{\tabcolsep}{1.5mm}
\begin{tabular}{lrccccccccccc}
\toprule  
\multicolumn{1}{c}{\multirow{2}{*}{\centering \textbf{Video encoder}}}  & {\textbf{MVBench}} & {\textbf{PerceptionTest } }  &{\textbf{LongVideoBench}} & {\textbf{MLVU}} & \textbf{VideoMME (\textit{w/o sub.})} & {\textbf{LVBench}}  \\ 
\cline{8-9}
 & Avg &Val&Val&M-Avg& Overall  & Avg  \\
\rowcolor{gray!10} 
Avg. Duration  & 16s& 23s  & 473s & 651s &  1010s &  4101s  \\
\midrule

 UMT-L &  73.2 & 75.6 & 64.2  & 74.5 & 64.0 &  48.4  \\

 InternVideo2-1B &  74.\textcolor{red}{(+1.1)} & 76.3\textcolor{red}{(+0.7)} & 64.5\textcolor{red}{(+0.3)} & 73.4\textcolor{teal}{(-1.1)} & 65.2\textcolor{red}{(+1.2)} & 48.7\textcolor{red}{(+0.3)} \\
\bottomrule
\end{tabular}
\end{adjustbox}
\caption{\textbf{Results with different video encoder.}}
\label{tab:internvideo2}
\vspace{-2mm}
\end{table*}
\subsection{Results with InternVideo2}
As shown in \cref{tab:internvideo2}, in addition to UMT~\cite{umt}, we also attempted to use the more powerful InternVideo2-1B~\cite{internvideo2} as the video encoder. As shown in Table 1, we found that a stronger video encoder can lead to better compressed representations.

\subsection{Results on Image Understanding Benchmarks} Our model is specifically designed for video understanding. However, according to the newly-evaluated results of image benchmarks, our model can still outperform the strong image-based MLLM, LLaVA-NeXT~\cite{llavanextvideo}, with significantly lower computational cost: MMMU (45.2 vs. 35.3), MME (1843.4 vs 1603.7).

\section{Implementation Details}
\subsection{Video-Language Connectors}
\begin{figure}
    \centering
    \includegraphics[width=1.0\linewidth]{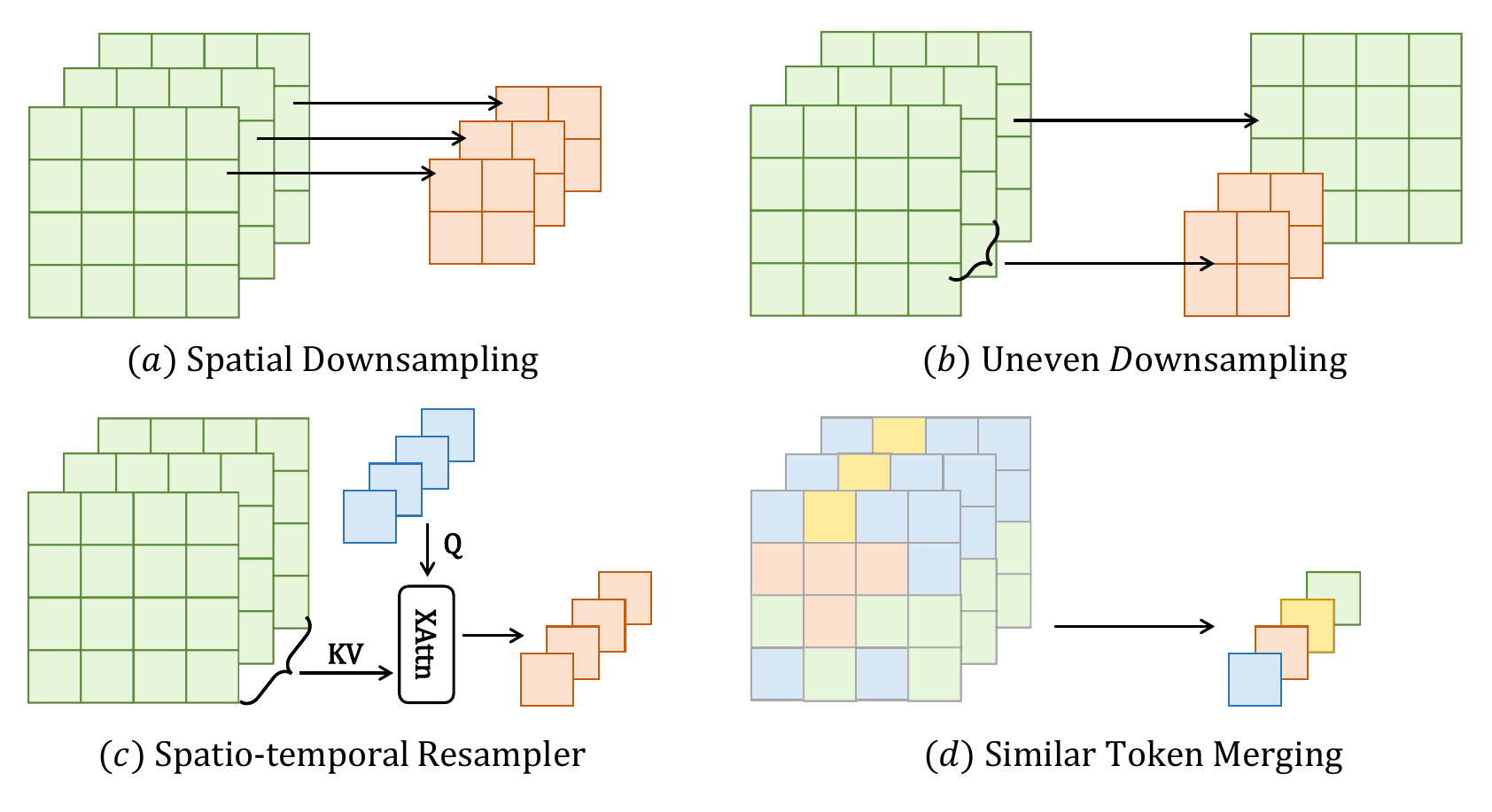}
    \caption{\textbf{Comparison of different connectors}.}
    \label{fig:compressor}
\end{figure}
As shown in \cref{fig:compressor}, we consider four popular token compression strategies to compress the features from video clips:
\begin{itemize}
    \item \textbf{\textit{Spatial Downsampling.}} Applying spatial operations (pooling~\cite{pllava}, interpolation~\cite{llavanextvideo}, and convolution (pixel shuffle)~\cite{internvl2}) to each video frame for downsampling has been demonstrated in previous work~\cite{pllava,videogpt+} as an effective method to reduce the number of video tokens. However, due to the lack of temporal interaction, this approach fails to leverage the relation between frames. We use pixel shuffle in our experiments.
    \item \textbf{\textit{Uneven Downsampling.}} Considering the similarities between adjacent frames, it is unnecessary to retain full details for every frame. We can apply down-sampling operations with different sizes across frames within a clip. Specifically, a lower down-sampling size is applied to the first frame, while higher down-sampling sizes are used for the remaining frames. Similar approaches have been validated in a recent study~\cite{uneven}. 
    \item \textbf{\textit{Spatio-Temporal Resampler.}} Using a learnable compressor, such as a Q-Former~\cite{videochat2} or a cross-attention layer, to compress spatiotemporal tokens. However, this approach requires a large amount of data for effective learning. In training, we observe that the Q-Former barely converges well in our setting. So in our ablations, we adopt a single-layer cross-attention instead.
    \item \textbf{\textit{Similar Token Merging.}} We directly merge similar tokens, using the ToMe~\cite{tome} approach.
\end{itemize}

\subsection{Training hyperparameters.} As shown in Table 1, the training details and hyperparameters for each stage of our VideoChat-Flash model are presented. 

\begin{table*}[htp]
    \centering
    \setlength{\tabcolsep}{12pt}
    \renewcommand{\arraystretch}{1.2}
    \resizebox{\textwidth}{!}{%
    \begin{tabular}{@{}ll|c|c|c|c@{}}
    \toprule
    & & \textbf{Stage-1}  & \textbf{Stage-2} & \textbf{Stage-3} &  \textbf{Stage-4} \\ 
    \midrule 
    \multirow{2}{*}{\rotatebox[origin=c]{90}{\footnotesize \textit{Vision}}}
    & \textbf{Resolution$\times$Num. frames}  & 224 & 224 $\times$8 & 224$\times$(64$\sim$512) & 224$\times$(64$\sim$512)  \\
    & \#Tokens & 16$\times$4  & 16$\times$8 & 16$\times$(64$\sim$512) & 16$\times$(64$\sim$512)    \\
    \midrule 
    \multirow{2}{*}{\rotatebox[origin=c]{90}{\footnotesize \textit{Data}}}
    & \textbf{Dataset} & Image \& Short Video & Image \& Short Video
    & (Multi)-Image \& Short/Long Video & (Multi)-Image \& Short/Long Video \\
    & \#Samples & 1M & 4M & 3.2M & 0.3M \\
    \midrule
    \multirow{2}{*}{\rotatebox[origin=c]{90}{\footnotesize \textit{Model}}}
    & \textbf{Trainable} & Projector & Full Model & Full Model & ViT\&Projector \\
    & 7.6B LLM & 20.0M & 7.9B & 7.9B & 0.3B\\
    \midrule 
    \multirow{4}{*}{\rotatebox[origin=c]{90}{\footnotesize \textit{Training}}}
    & \textbf{Batch Size} & 512 & 256 & 256  & 256 \\
    & \textbf{LR} of \textit{vision encoder} & 1$\times 10^{-3}$ & 2$\times 10^{-6}$ & 2$\times 10^{-6}$  & 2$\times 10^{-6}$ \\    
    & \textbf{LR}\textit{ of connector \& LLM} & 1$\times 10^{-3}$ & 1 $\times 10^{-5}$ & 1 $\times 10^{-5}$  & 1 $\times 10^{-5}$  \\
    & \textbf{Epoch} & 1 & 1 & 1 & 1 \\
    \bottomrule
    \end{tabular}
    }
    \vspace{1mm}
    \caption{\textbf{Training details of each training stage for the VideoChat-Flash-7B model}}
    \vspace{-5mm}
    \label{tab:training_strategy}
\end{table*}

\subsection{Training Data}
\paragraph{Stage 1: Video-Language Alignment.} In this stage,  we use 558k image-text pairs from LCS-558K~\cite{llava} and 481k short video-text pairs from S-MiT~\cite{smit}.

\paragraph{Stage 2: Short Video Pre-training.} To enhance the model's understanding of visual concepts, we conduct visual pre-training using 3.5 million images and 2.5 million short video-text pairs.

\begin{itemize}
    \item \textbf{\textit{Video Description Data.}} We utilize the video description data recaptioned with VideoChat2~\cite{videochat2} from WebVid2M~\cite{webvid}.
    \item \textbf{\textit{Detailed Video Description Data}}. We employ the 323k detailed video description data recaptioned with Gemini~\cite{gemini} from WebVid~\cite{webvid} and Kinetics~\cite{kinetics}, as in previous work~\cite{sharegemini}. 
    \item \textbf{\textit{Detailed Image Description Data.}} We use the 3.5 million detailed image description data recaptioned with LLava-NeXT-34B~\cite{llavanextvideo} from the following datasets: COCO118K, BLIP558K, and CC3M, as provided by previous work~\cite{llavaonevision}.
    \item \textbf{\textit{Text Data.}} To enhance the model's language understanding capabilities, we incorporate 143K samples from the Evo-Instruct dataset~\cite{allava}.
\end{itemize}

\paragraph{Stage 3: Joint Short \& Long Video Instruction tuning.} To enable the model to handle a wide variety of video tasks, we collect 3.5 million instruction fine-tuning samples, including 1.1M images, 1.7M short videos (under 60 seconds), and 0.7M long videos (60$\sim$3600 seconds). 

\begin{itemize}
    \item \textbf{\textit{Image Instruction data.}} We primarily utilized single-image instruction data from LLava-NeXT~\cite{llavanextvideo}, Allava~\cite{allava}, and ShareGPT4-o~\cite{internvl2,internvideo2}. Additionally, we incorporated multi-image data provided by LLaVA-Interleave~\cite{llavainterleave}. 
    \item  \textbf{\textit{Short Video Instruction data.}} We primarily utilized short video data from VideoChat2~\cite{videochat2} and InternVideo2~\cite{internvideo2} for instruction fine-tuning. Additionally, we incorporated data annotated with GPT4-o from previous works, including ShareGPT4o~\cite{internvl2,internvideo2}, VideoChatGPT-Plus~\cite{videogpt+}, LLaVA-Video-178K~\cite{llavavideo} and LLava-Hound~\cite{llavahound}.
    \item \textbf{\textit{Long Video Instruction data.}} We primarily utilized long video instruction data from MoiveChat~\cite{moviechat}, Vript~\cite{vript} and our LongVid.
    
\end{itemize}

\section{Dataset Details of LongVid}

The videos of LongVid are curated from 4 open-source video datasets: Ego4D~\cite{ego4d}, HowTo100M~\cite{howto100m}, HD-VILA~\cite{hdvila}, and MiraData~\cite{miradata}. We provide details of the data construction pipeline for each dataset as follows.  

\subsection{Ego4D}  

For ego-centric videos, we adopt 3,662 long videos from the Ego4d~\cite{ego4d} and leverage Ego4DHcap~\cite{videorecap} as the corresponding captions. Ego4DHcap gives hierarchical captions for short, medium, and long video segments. For the short video captioning task, we directly utilize these captions, while for the dense caption task, we concatenate captions in the lower level to form a dense one. For example, we merge all short video captions in a medium video segment to create a dense medium-level one, and the dense caption of long video segments can be formed by concatenating multiple medium-level video captions. 

We also build event relation recognition and temporal grounding tasks based on captions of short video segments. For the event relation recognition task, models are required to choose the right order of an event sequence. Since we find the captions of short videos are highly concise and event-oriented, we use them as the event labels and serially put the short captions in a medium-level video segment as the ground-truth event relationship. For the temporal grounding task, we use the short video captions with the corresponding timestamps as the ground-truth, and randomly select other timestamps in the current medium video segments as the false options. 

\subsection{MiraData}

MiraData~\cite{miradata} provides multi-level captions for large-scale minute-level movie segments. Apart from short and dense captions that are used for short and dense video captioning tasks, it also provides multiple fine-grained captions that focus on various specific perspectives, such as the main object, background, camera movements, and video style. We use an open-source LLM (Qwen-72b~\cite{qwen}) to extract the event and background labels from the main object and background captions, respectively, and we put the labels of a long video in the right order as the ground truth of the event/background relation recognition task. For the temporal grounding task, we use the event label with the corresponding timestamp as the ground-truth option.



\subsection{HowTo100M}

HowTo100M~\cite{howto100m} includes more than 1 million long-duration how-to videos. We adopt HowToInterlink7M~\cite{cosmo}, a video captioning dataset that provides refined interleaved video captions of HowTo100M videos as short and dense video captions. For the event relationship recognition and temporal grounding tasks, we use HTStep~\cite{htstep}, a large-scale dataset containing temporal annotations of instructional steps in HowTo100M videos.

\subsection{HD-VILA}

While previous datasets focus on long videos in specific domains, we also select part of the videos from HD-VILA~\cite{hdvila}, a large-scale video dataset that includes various in-the-wild videos. We argue that adding these videos into training could enhance the model's ability to process long videos in some uncommon domains. For HD-VILA videos, we adopt the captions of Panda-70M~\cite{panda70m}. Specifically, we filter consecutive video segments that can be re-constructed into more than 60s long videos from the 10M training subset and utilize these captions as the video short/dense captioning and temporal grounding tasks. The event labels are also extracted from these captions in the same way as MiraData~\cite{miradata}.

\section{Qualitative Results}

\begin{figure*}[!ht]
    \centering
    \includegraphics[width=0.9\linewidth]{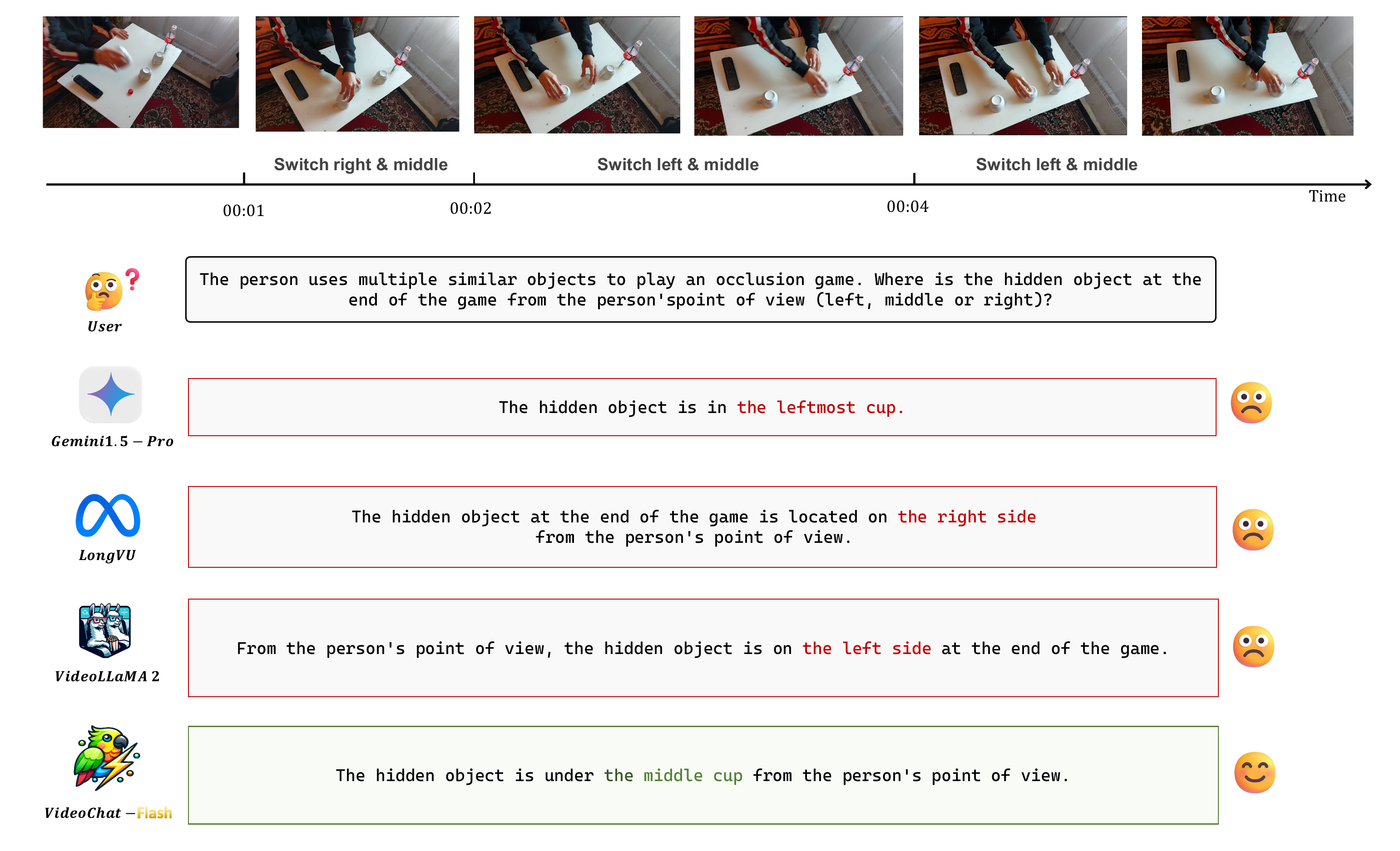}
    \caption{\textbf{Fine-grained Understanding of Short Videos: Fast Motion.} By adopting a dense sampling strategy for short videos, our model effectively captures fast motion within the video, enabling it to accurately determine the final position of the object under the cup.}
    \label{fig:demofastmotion}
\end{figure*}

\begin{figure*}[!ht]
    \centering
    \includegraphics[width=0.9\linewidth]{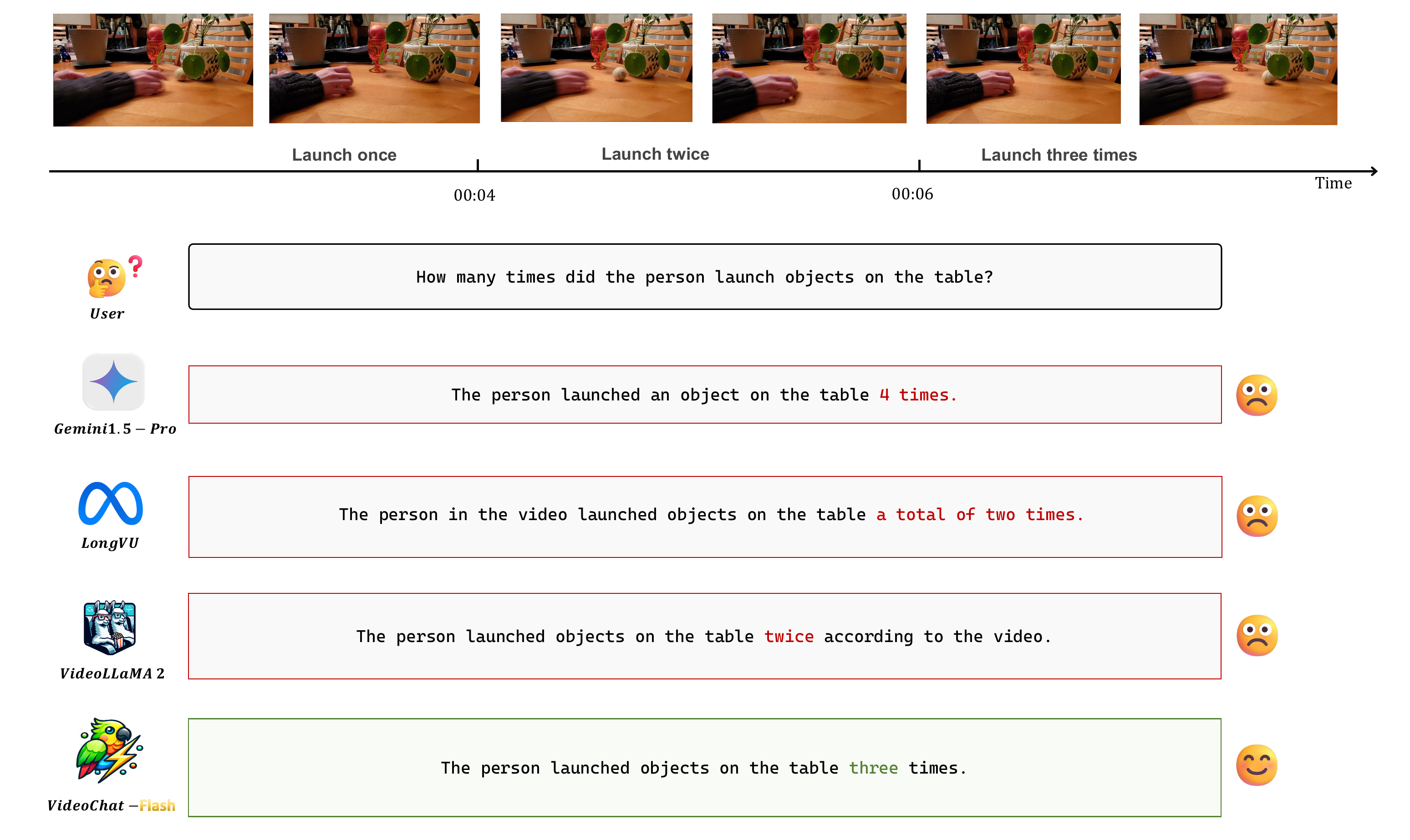}
    \caption{\textbf{Fine-grained Understanding of Short Videos: Action Count.} Our model can accurately capture actions in short videos while also recording their occurrence frequency.}
    \label{fig:democount}
\end{figure*}

\begin{figure*}[!ht]
    \centering
    \includegraphics[width=0.85\linewidth]{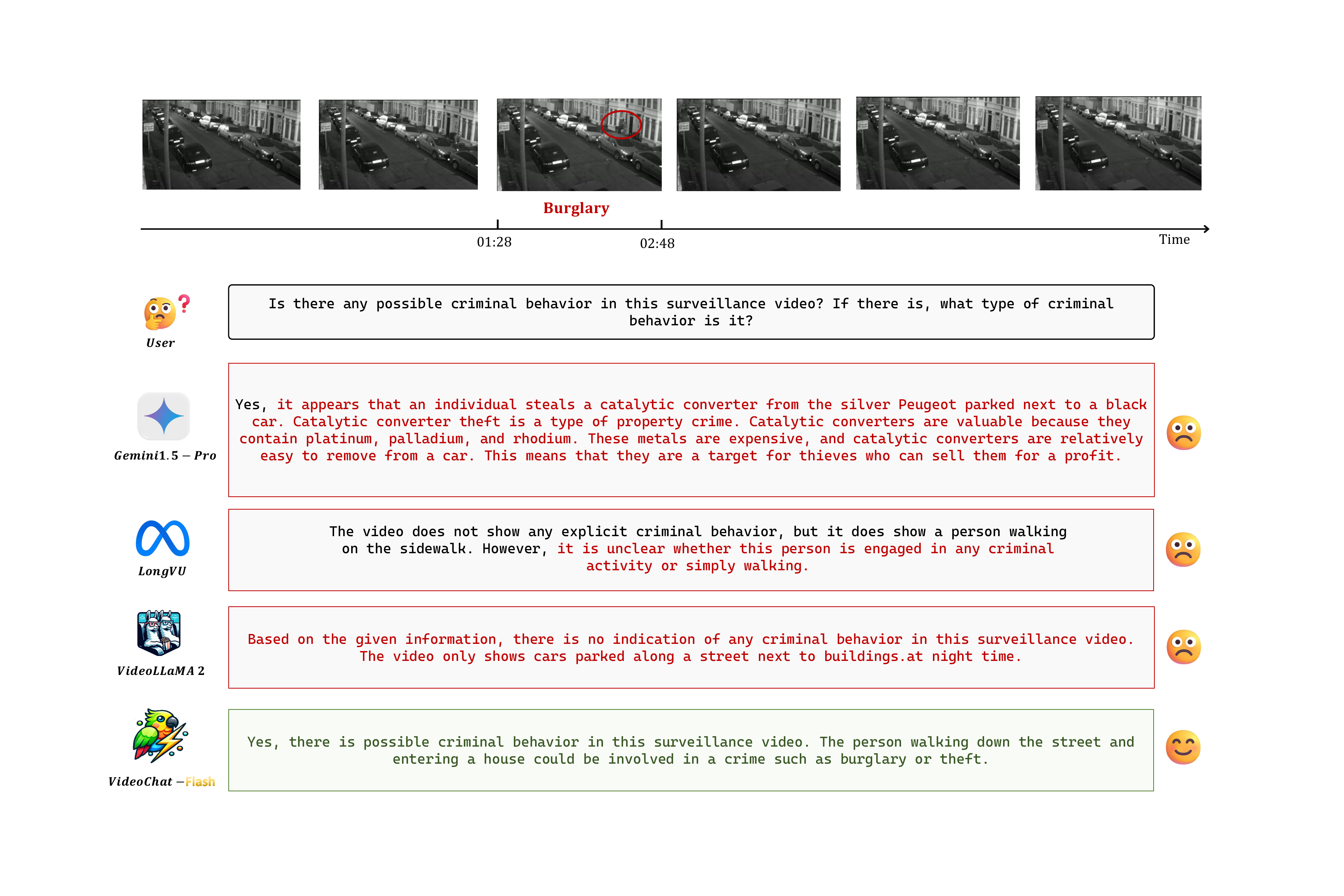}
    \caption{\textbf{Long video understanding: Anomaly Detection In Surveillance Videos.} Our model can detect anomalous behaviors in surveillance videos and provide corresponding inferences.}
    \label{fig:demoanoma}
\end{figure*}

\begin{figure*}[!ht]
    \centering
    \includegraphics[width=0.9\linewidth]{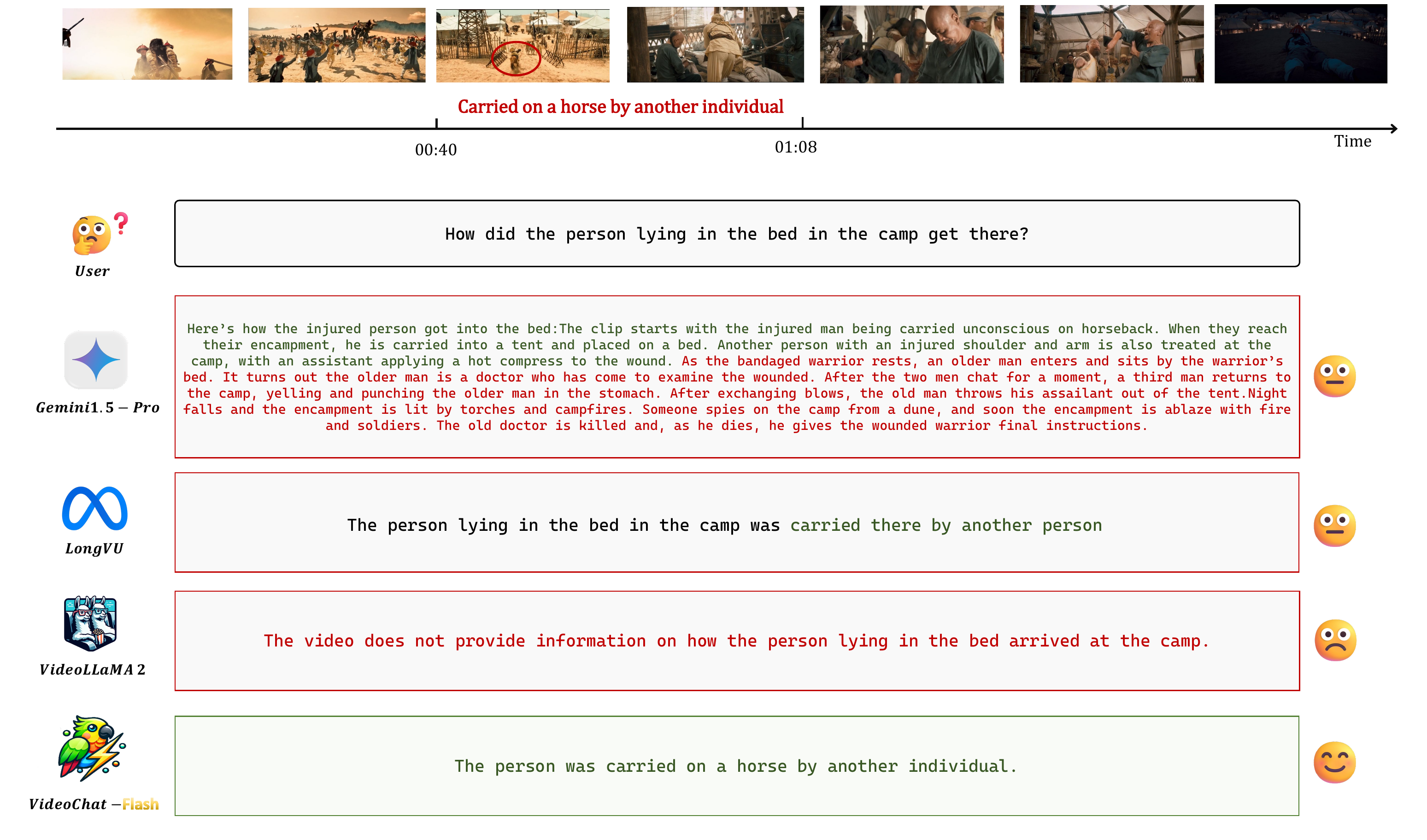}
    \caption{\textbf{Long video understanding: Moive Understanding.} Our model can understand the plot of a movie and retain detailed visuals.}
    \label{fig:demomovie}
\end{figure*}

We perform qualitative comparisons of our model with the proprietary model Gemini-1.5 Pro~\cite{gemini}\footnote{We use the newest Gemini-1.5 Pro-002 for evaluation.} and the open-source LongVU~\cite{longvu} and VideoLLaMA2~\cite{videollama2} across three tasks: fine-grained understanding of short videos (~\cref{fig:demofastmotion,fig:democount}) and long video understanding (~\cref{fig:demoanoma,fig:demomovie}).


\end{document}